\documentclass{article}
\pdfoutput=1

\usepackage{iclr2024_conference,times}
\iclrfinalcopy 

\PassOptionsToPackage{compress,sort}{natbib}

\usepackage{selectp}

\usepackage[utf8]{inputenc} 
\usepackage[T1]{fontenc}    
\usepackage[hidelinks, colorlinks=true,allcolors=teal]{hyperref}       
\usepackage{url}            
\usepackage{booktabs}       
\usepackage{amsfonts}       
\usepackage{nicefrac}       
\usepackage{microtype}      
\usepackage{xcolor}         
\usepackage{graphicx}

\usepackage{amsmath}
\usepackage{amssymb}
\usepackage{mathtools}
\usepackage{amsthm}
\usepackage[font=small,labelfont=bf,tableposition=top]{caption}
\DeclareCaptionLabelFormat{andtable}{#1~#2  \&  \tablename~\thetable}

\usepackage[capitalize]{cleveref}

\theoremstyle{plain}
\newtheorem{theorem}{Theorem}[section]
\newtheorem{proposition}[theorem]{Proposition}

\newtheorem{corollary}[theorem]{Corollary}
\theoremstyle{definition}
\newtheorem{definition}[theorem]{Definition}

\theoremstyle{remark}

\usepackage{algorithm}
\usepackage[noend]{algpseudocode}
\usepackage[subtle]{savetrees}
\usepackage[compact]{titlesec}
\graphicspath{ {./fig/} }
\usepackage{subcaption}
\usepackage{multirow}
\usepackage{hhline}
\usepackage{enumitem}
\setlist{nolistsep}
\global\long\def\bR{\mathbb{R}}
\usepackage{placeins}
\usepackage[acronym,nohypertypes={acronym,notation}]{glossaries}
\glsdisablehyper
\newacronym{dnn}{DNN}{Deep Neural Network}
\newacronym{cnn}{CNN}{Convolutional Neural Network}
\newacronym{nn}{NN}{Neural Network}
\newacronym{ann}{ANN}{Artificial Neural Network}
\newacronym{snn}{SNN}{Sparse Neural Network}
\newacronym{lt}{LT}{Lottery Ticket}
\newacronym{ntk}{NTK}{the Neural Tangent Kernel}
\newacronym{lth}{LTH}{Lottery Ticket Hypothesis}
\newacronym{nlp}{NLP}{Natural Language Processing}
\newacronym{dst}{DST}{Dynamic Sparse Training}
\newacronym{gpu}{GPU}{Graphics Processing Unit}
\newacronym{pi}{PI}{Primary Investigator}
\newacronym{ai}{AI}{Artificial Intelligence}
\newacronym{ml}{ML}{Machine Learning}
\newacronym{cv}{CV}{Computer Vision}
\newacronym{neurips}{NeurIPS}{Neural Information Processing Systems}
\newacronym{iclr}{ICLR}{the International Conference on Learning Representations}
\newacronym{cvpr}{CVPR}{the IEEE/CVF Conference on Computer Vision and Pattern Recognition}
\newacronym{flops}{FLOPs}{Floating Point Operations}
\newacronym{relu}{ReLU}{ReLU}
\newacronym{srigl}{SRigL}{Structured RigL}
\newacronym{rigl}{RigL}{Rigging the Lottery Ticket}
\newacronym{set}{SET}{Sparse Evolutionary Training}
\newacronym{erk}{ERK}{Erd\H{o}s-R\'enyi-Kernel}
\newacronym{ste}{STE}{Straight-Through Estimator}
\newacronym{srste}{SR-STE}{Sparse-Refined Straight-Through Estimator}
\newacronym{ilsvrc}{ILSVRC-12}{2012 ImageNet Large Scale Visual Recognition Challenge}
\newacronym{ast}{AST}{Alternating Sparse Training}
\newacronym{deepr}{DeepR}{Deep Rewiring}
\newacronym{snfs}{SNFS}{Sparse Networks from Scratch}
\newacronym{dsr}{DSR}{Dynamic Sparse Reparameterization}
\newacronym{topkast}{Top-KAST}{Top-K Always Sparse Training}
\newacronym{mest}{MEST}{Memory-Economic Sparse Training}
\newacronym{dsb}{DSB}{Dynamic Shuffled Block}
\newacronym{dynsparse}{DynSparse}{Dynamic Sparsity}
\newacronym{vit}{ViT-B/16}{Vision Transformer}
\newacronym{cpu}{CPU}{Central Processing Unit}
\newacronym{csr}{CSR}{Compressed Sparse Row}
\newacronym{itop}{ITOP}{In Time Overparameterization Rate}

\usepackage{threeparttable} 
\usepackage{appendix}
\title{Dynamic Sparse Training\newline with Structured Sparsity}

\author{%
  Mike Lasby$^{1}$, Anna Golubeva$^{2,3}$, Utku Evci$^{4}$, Mihai Nica$^{5,6}$, Yani A. Ioannou$^{1}$\\
  $^1$University of Calgary, $^2$Massachusetts Institute of Technology, $^3$IAIFI\\
  $^4$Google DeepMind, $^5$University of Guelph, $^6$Vector Institute for AI
\thanks{
  \href{mailto:mklasby@ucalgary.ca}{\{mklasby,yani.ioannou\}@ucalgary.ca}, %
  \href{mailto:golubeva@mit.edu}{golubeva@mit.edu}, %
  \href{mailto:evcu@google.com}{evcu@google.com}, %
  \href{mailto:nicam@uoguelph.ca}{nicam@uoguelph.ca}%
}
}
\begin{document}
\maketitle
\begin{abstract}
\Gls{dst} methods achieve state-of-the-art results in sparse neural network training, matching the generalization of dense models while enabling sparse training and inference. Although the resulting models are highly sparse and theoretically less computationally expensive, achieving speedups with unstructured sparsity on real-world hardware is challenging. 
In this work, we propose a sparse-to-sparse \gls{dst} method, \gls{srigl}, to learn a variant of fine-grained \emph{structured} N:M sparsity by imposing a \emph{constant fan-in} constraint.
Using our empirical analysis of existing \gls{dst} methods at high sparsity, we additionally employ a neuron ablation method which enables \gls{srigl} to achieve state-of-the-art sparse-to-sparse structured \gls{dst} performance on a variety of \gls{nn} architectures. %
Using a 90\% sparse linear layer, we demonstrate a real-world acceleration of 3.4$\times$/2.5$\times$ on CPU for \emph{online inference} and 1.7$\times$/13.0$\times$ on GPU for inference with a batch size of 256 when compared to equivalent dense/unstructured (CSR) sparse layers, respectively.%
\newcommand\blfootnote[1]{%
  \begingroup
  \renewcommand\thefootnote{}\footnote{#1}%
  \addtocounter{footnote}{-1}%
  \endgroup
}
\blfootnote{Our source code is available \href{https://github.com/calgaryml/condensed-sparsity}{here}.}
\end{abstract}
%
\glsresetall
\glsunset{rigl} 
\section{Introduction}
\label{sec:intro}
\Gls{dst} methods such as \gls{rigl}~\citep{evci_rigging_2021} are the state-of-the-art in sparse training methods for \glspl{dnn}. \gls{dst} methods typically learn \emph{unstructured} masks resulting in 85--95\% fewer weights than dense models, while maintaining dense-like generalization and typically outperforming masks found via pruning. Furthermore, sparse-to-sparse \gls{dst} algorithms are capable of employing sparsity \emph{both during training and inference}, unlike pruning and dense-to-sparse \gls{dst} methods such as SR-STE~\citep{zhou2021learning} which only exploit sparsity at inference time. 

\begin{figure}[tb]
    \begin{subfigure}{0.64\linewidth}
    \centering
    \includegraphics[width=\linewidth]{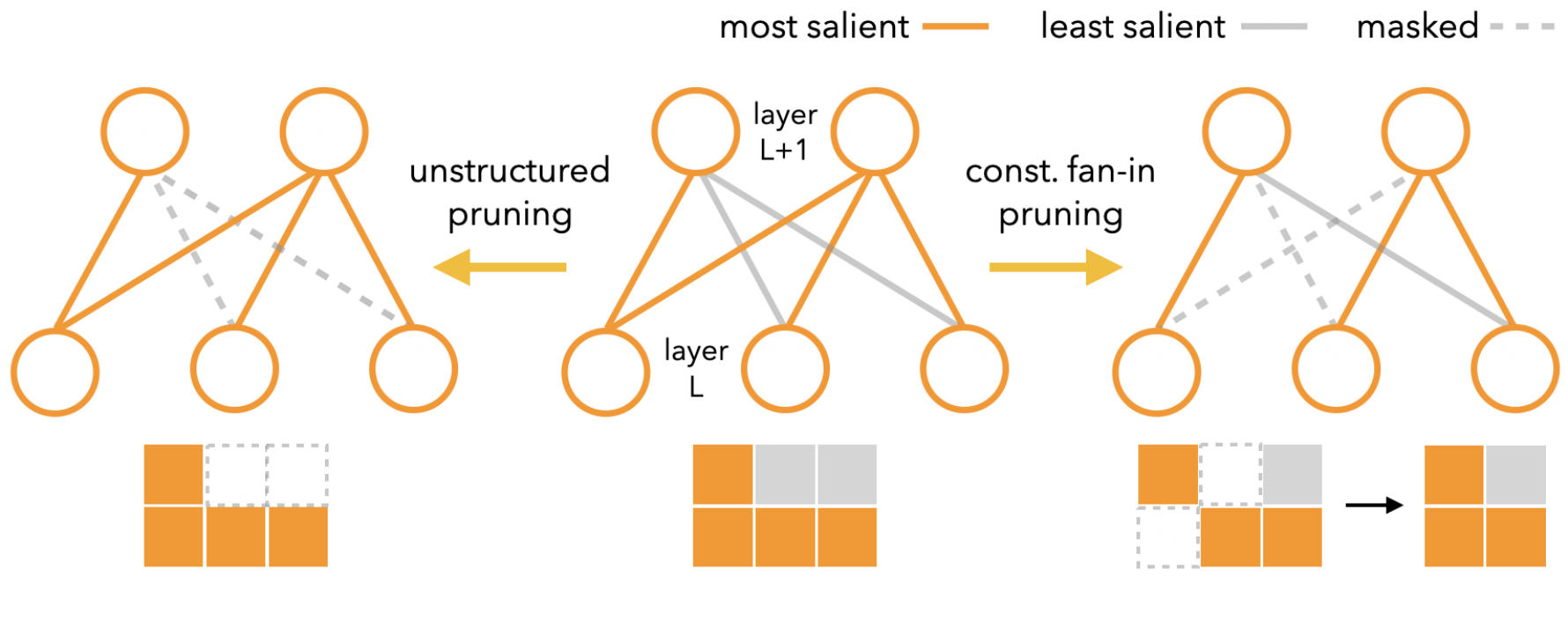}
    \caption{Constant fan-in pruning v.s.\ unstructured pruning.}\label{fig:constfanin}
    \end{subfigure}
    \hfill
    \begin{subfigure}{0.34\linewidth}
    \centering
    \includegraphics[width=\linewidth]{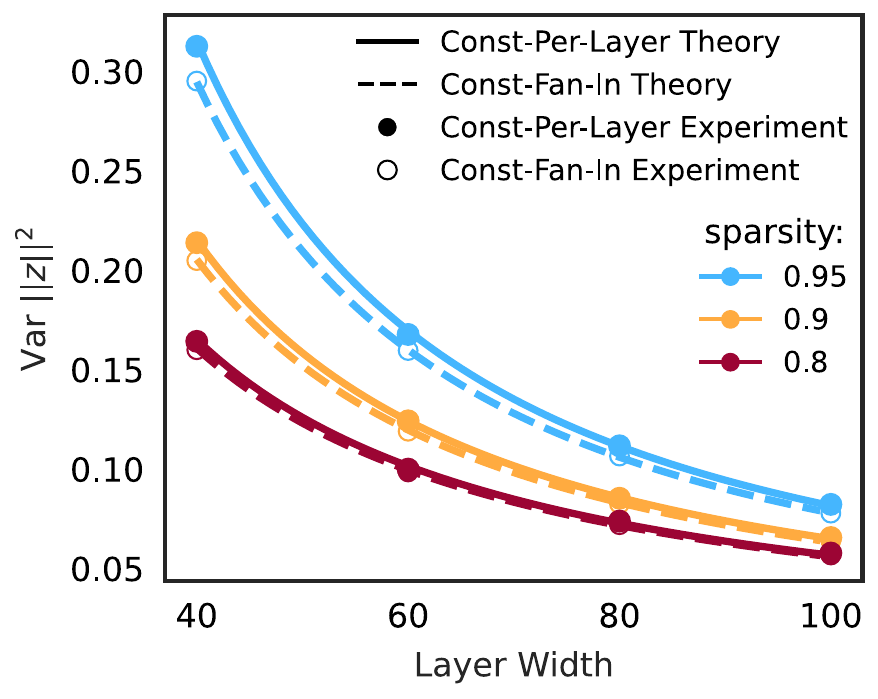}
    \caption{Output-norm variance analysis.}\label{fig:theory}
    \end{subfigure}
    \caption{(\subref{fig:constfanin}) \textbf{Constant fan-in} pruning keeps the most salient weights \emph{per neuron}, while unstructured pruning keeps the most salient weights \emph{per layer}. A constant fan-in weight matrix has the same number of non-zero elements (here 2) per column allowing condensed representation. While pruning may remove salient weights affecting generalization, with \acrshort{srigl} structure and weights are learned concurrently. (\subref{fig:theory}) \textbf{Output-norm variance}: Theoretical predictions and simulation results (see \cref{sec:outputnormvariance}) demonstrating that sparse layers with constant fan-in have consistently smaller output-norm variance than layers with the same sparsity but w/o the constant fan-in constraint. 
    }
\end{figure}

While models trained with \gls{dst} methods are highly sparse and enable a large reduction in \gls{flops} in theory, realizing these speedups on hardware is challenging when the sparsity pattern is unstructured. Even considering recent advances in accelerating unstructured \glspl{snn}~\citep{gale2020sparse,elsen2020cvpr,ji_fscnn_2022}, structured sparsity realizes much stronger acceleration on real-world hardware. On the other hand, structured sparse pruning often removes salient weights, resulting in worse generalization than comparable unstructured \glspl{snn} for the same sparsity level (\cref{fig:constfanin}). 
Our work presents a best-of-both-worlds approach: we exploit the \gls{dst} framework to learn \emph{both} a highly-sparse \emph{and} structured representation while maintaining generalization performance.
In summary, our work makes the following contributions:
\begin{enumerate}[noitemsep]
    \item We propose a novel sparse-to-sparse \gls{dst} method, \gls{srigl}, based on \gls{rigl}~\citep{evci_rigging_2021}. \Gls{srigl} learns a \gls{snn} with constant fan-in fine-grained structured sparsity (\cref{fig:constfanin}) while maintaining generalization comparable with \gls{rigl} up to a high sparsity level (99\%) for a variety of network architectures.
    This structure is a particular case of ``N:M sparsity'' which requires $N$ out of $M$ consecutive weights to be non-zero~\citep{mishra_accelerating_2021}.
    \item Our empirical analysis shows \gls{rigl}, at sparsity levels > 90\%, ablates whole neurons. By allowing neuron ablation in \gls{srigl}, we match  \gls{rigl} generalization even in this high-sparsity regime.
    \item We enable neuron ablation in \gls{srigl} across all sparsity regimes. We find this structured sparsity is complementary to the constant fan-in sparsity in improving real-world inference timings while maintaining generalization comparable to unstructured \gls{dst} methods.
    \item We demonstrate that constant fan-in sparsity enables a compact representation that is not only parameter- and memory-efficient, but also amenable to real-world acceleration. We observe significantly reduced real-world timings for online inference using our CPU-based PyTorch implementation and for batched inference using a GPU-based implementation from \citet{schultheis_towards_2023} over dense and unstructured baselines.  
\end{enumerate}
\section{Related work}
\paragraph{Dynamic sparse training}
Unlike with pruning, where weights are typically pruned after the dense network was trained~\citep{han2015learning,deepcompression}, or at initialization~\citep{grasp}, \gls{dst} methods learn the sparse connectivity during training by periodically adding and removing weights based on various saliency criteria. For instance, \Gls{set}~\citep{mocanu_scalable_2018} removes weights with the smallest magnitude and adds weights randomly; similarly, \gls{rigl}~\citep{evci_rigging_2021} prunes weights with the smallest magnitude and regrows weights that have large-magnitude gradients.
\citet{liu_we_2021} further improved the original \gls{rigl} results by increasing the extent of the parameter space explored by modifying the sparse connectivity update schedule and drop rate. 

Many recent works have examined the effect of different grow and prune saliency criteria on unstructured \gls{dst} approaches, including \gls{set}, \gls{deepr} \citep{bellec_deep_2018}, \gls{snfs} \citep{dettmers_sparse_2019}, \gls{dsr} \citep{mostafa_parameter_2019}, \gls{topkast} \citep{jayakumar_top-kast_2020}, and \gls{mest} \citep{yuan_mest_2021}. In \cref{sec:results}, we compare \gls{srigl} to several of these methods. While the above-noted \gls{dst} methods are highly effective at finding \glspl{snn} which reduce theoretical inference cost, they result in unstructured \glspl{snn} which are difficult to accelerate in practice on common hardware architectures.

In a contemporaneous work, \citet{yin_dynamic_2023} also identified the existence of \emph{sparse amenable channels} in existing unstructured \gls{dst} algorithms. Their method, Chase, achieves state-of-the-art generalization performance by including a soft memory bound similar to \citet{yuan2021mest} and calculating the saliency of parameters based on global instead of layer-wise statistics.
Chase requires that the structured sparsity level be set prior to training. In contrast, \gls{srigl} dynamically learns to ablate channels based on the number of remaining weights that are considered salient. 
\paragraph{Accelerating unstructured sparse neural networks}
\Citet{elsen2020cvpr} proposed a method for accelerating unstructured \glspl{snn} based on one-dimensional tiling of non-zero elements, which demonstrated significant speedups on both \gls{cpu}~\citep{elsen2020cvpr} and \gls{gpu}~\citep{gale2020sparse}. However, like most approaches to accelerating unstructured \glspl{snn}, this method relies on imposing structure on an existing sparse weight matrix \emph{after training}. Our method can be considered a way of adding structure to \glspl{snn} \emph{during training}, allowing the model to maximally utilize non-zero weights since structure and weights are learned concurrently.

DeepSparse Engine~\citep{neural_magic_neuralmagicdeepsparse_2021} accelerates inference of unstructured sparse networks on \gls{cpu} by applying several innovations. In \cref{sec:deepsparse}, we compare our timings with \gls{srigl} to the DeepSparse Engine.

\paragraph{Learning block structured sparsity from scratch}
Block sparsity is a particular type of structured sparsity in which blocks of non-zero weights are grouped together in arrangements that reduce the memory overhead required to store the indices of the non-zero weights. Blocks can be generated out of contiguous weights in 1D (sometimes called tiles) or 2D or by utilizing a fixed number of non-zero weights per row or column group in the case of block-balanced sparsity~\citep{hoefler_sparsity_2021}. 
Spurred by the success of \gls{dst} in learning unstructured sparse models, recent works have attempted to apply \gls{dst} principles to learn block-structured sparsity. \Citet{jiang_exposing_2022} introduced a novel block-aware \gls{dst} algorithm known as \gls{dsb}. \gls{dsb} reshuffles non-zero weights into a block sparsity pattern after sparse connectivity updates, thereby improving memory access efficiency. Wall-clock speed-ups of up to 4$\times$ were reported with this method; however, generalization performance was reduced compared to \gls{rigl} at comparable sparsities. 
\Citet{dietrich_towards_2022} applied a modified variant of \gls{rigl} to BERT models \citep{devlin_bert_2019}. The resulting method is capable of learning models with block-structured sparsity. 
\paragraph{Learning N:M structured sparsity from scratch} 
N:M sparsity is a specific form of block-balanced sparsity in which 1D blocks with $M$ contiguous elements contain exactly $N$ non-zero elements. N:M sparsity is particularly amenable to acceleration and several attempts have been made to train models with N:M fine-grained structure using \gls{dst} methods. 

\Citet{yang_get_2022} extended the \gls{dst} method proposed by \citet{liu_sparse_2021} to train multiple sparse sub-networks sampled from a single dense super-network. Their proposed method, \gls{ast}, switches the network topology between sparse sub-networks after each mini-batch during training. \citet{yang_get_2022} demonstrated state-of-the-art performance on several typical sparse training benchmarks.
However, the dense model weights and gradients are required throughout the majority of training, greatly increasing the overall compute and storage requirements.
While \gls{ast} demonstrated a tantalizing possibility of training multiple sparse sub-networks within a single training loop, the gradual dense-to-sparse training paradigm used by \citep{liu_sparse_2021} is not directly comparable to \gls{rigl} or other similar end-to-end sparse \gls{dst} methods.

\Citet{zhou2021learning} explored how N:M sparsity can be achieved during training using magnitude-based pruning during the forward pass and a \gls{ste}~\citep{bengio2013ste} on the backward pass. In their method, the dense network weights are projected into a sparse network during each training iteration. The sparse network is obtained by selecting the top-N out of every M contiguous weights and \Gls{ste} is used to propagate the approximated gradients through the projection function.
A regularization term is applied to the gradients of pruned weights to reduce instabilities during training. Their approach --- \gls{srste} --- was applied to networks with N:M ratios of 1:4, 2:4, 2:8, 4:8, 1:16.

Although \gls{srste} utilizes sparse operations in the forward pass and can find sparse models optimized for inference, it does not reduce the training cost significantly. Specifically, \gls{srste} training requires (1) storing original parameters in their dense format, and (2) calculating dense gradients during each training iteration. This makes \gls{srste} training as expensive as the original dense training in terms of memory and compute cost\footnote{To be precise, \gls{srste} can use some sparse operations and reduce training cost up to two thirds of the original dense training. However this is still far from fully sparse acceleration for training.}. On the other hand, \Gls{dst} methods such as \gls{rigl}, and our proposed method \gls{srigl}, are capable of end-to-end sparse training and use sparse parameters and gradients throughout training.
\paragraph{Accelerating fine-grained N:M structured sparsity} 
\Citet{NvidiaA100,mishra_accelerating_2021} introduced the Ampere Tensor Core \gls{gpu} architecture (e.g.\ A100 \glspl{gpu}) and proposed the 2:4 fine-grained structured sparsity scheme that enables~\gls{snn}s to be accelerated on this hardware \textit{at inference time}. This scheme places a constraint on the allowed sparsity pattern: For every contiguous array of four weights, two are pruned, yielding a 50\%-sparse net. The resulting regular structure of the weight matrix allows one to compress it efficiently and to reduce memory storage and bandwidth by operating on the nonzero weights only. Since the focus is on acceleration at inference time, the authors proposed to use the standard method of magnitude-based pruning post training to achieve the 2:4 sparsity. Importantly, this work considered exclusively the 2:4 ratio; other N:M ratios cannot be accelerated on Ampere \glspl{gpu}.
\paragraph{Constant fan-in N:M structured sparsity}\label{sec:constfaninbackground}
The constant fan-in constraint represents a special case of N:M sparsity where $N$ is the number of non-zero weights per neuron and $M$ is the dense fan-in for each neuron within a given layer. While commodity hardware acceleration currently exists only for 2:4 sparsity on Nvidia's Ampere and later architectures~\citep{mishra_accelerating_2021}, a constant fan-in constraint can also take advantage of the efficient memory access and throughput increase that N:M sparsity yields, as recently demonstrated by \citet{schultheis_towards_2023}. 
Constant fan-in sparsity has several attributes which differentiate it from N:M sparsity:
\begin{itemize}[itemsep=0.2em, leftmargin=2em]
    \item Constant fan-in sparsity is more flexible than N:M sparsity, enabling arbitrary global sparsity values to be applied to the mode whereas N:M sparsity is limited to specific sparsity ratios.
    \item With the constant fan-in constraint, per-layer sparsity distributions such as \gls{erk} can be applied to the model. The \gls{erk} distribution has been demonstrated to outperform uniform sparsity distributions by reallocating parameters to layers with fewer parameters \citep{mocanu_scalable_2018, evci_rigging_2021}. In contrast, N:M sparsity can only be applied with a uniform sparsity distribution. 
    \item Hardware support for acceleration of N:M sparsity is currently limited to 2:4 sparsity on Nvidia \glspl{gpu}, offering a modest acceleration on the order of $\times$2. In contrast, the potential promise of highly sparse models (>=90\% sparsity)  to be $\times$10 faster than an equivalent dense model. As we demonstrate in \cref{sec:accel,sec:timingsdetails}, our condensed sparse representation with constant fan-in sparsity can achieve significant acceleration over a wide range of sparsities even without specialized hardware.  
\end{itemize}
\paragraph{Online inference}
In many applications, \glspl{dnn} are used in an \emph{online} manner, i.e.\ by using only single inputs and not batches of inputs. Online inference is common in real-time and latency-sensitive applications, or applications without significant numbers of simultaneous requests allowing batching. Online inference, especially for real-time applications, does not typically benefit from accelerators such as \glspl{gpu} that require host to device transfers, since the cost of the transfer itself often negates any benefit in compute. Accelerating online inference workloads remains an open research problem, with many systems engineering solutions proposed to achieve acceleration \citep{kumar_accelerating_2019,li_edge_2020,wang_merlin_2022,wu_irina_2020}. Our condensed representation CPU implementation, which exploits both structured and constant fan-in sparsity, offers a complimentary, orthogonal solution to these engineered solutions by directly accelerating model inference for single samples. 
\section{Method}
Our goal in this work is to introduce structural constraints on the sparse mask learned by \gls{rigl}, in order to make it more amenable to acceleration at inference time while not affecting \gls{rigl}'s generalization performance. We first performed a theoretical analysis to explore the effect of various sparsity distributions with different degrees of structural constraints on the training dynamics of \glspl{snn}, detailed in \cref{fig:constfanin,sec:outputnormvariance}. Based on this analysis, we did not find any evidence to suggest that the constant fan-in constraint would impair \gls{snn} training dynamics and performance, motivating the use of constant fan-in sparsity in our method outlined in \cref{sec:sriglmethod}.
\subsection{Structured \protect\glsentrytext{rigl}}\label{sec:sriglmethod}
As motivated by \cref{sec:outputnormvariance}, we propose to enforce the constant-fan-in constraint within a sparse-to-sparse \gls{dst} method to learn structured sparse connectivity from scratch. Specifically, we use \gls{rigl} by~\citet{evci_rigging_2021}, which can obtain highly sparse networks with generalization performance comparable to their dense baselines. 
\begin{figure}[tbp]
    \centering
    \includegraphics[width=\linewidth]{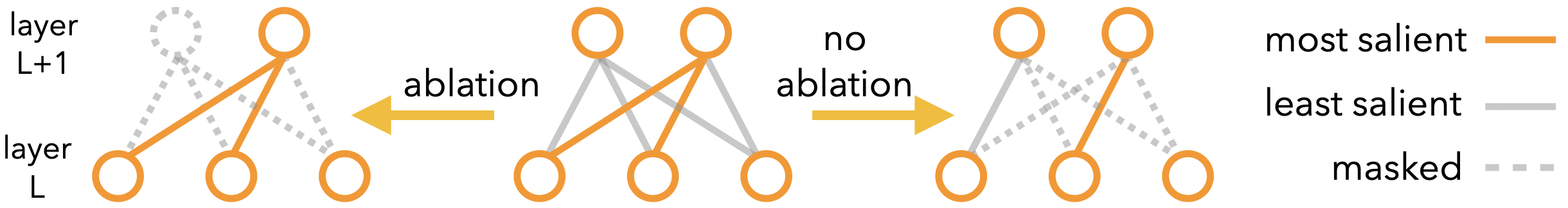}
    \caption{\textbf{Neuron ablation.} At sparsity levels over 90\%, \gls{rigl} learns to completely mask (ablate) a large number of neurons within each layer, effectively reducing layer width. Imposing a constant fan-in constraint requires all neurons to have the same number of (non-pruned) incoming weights and therefore inhibits ablation, which results in worse generalization performance than \gls{rigl}. Allowing \gls{srigl} to ablate neurons  restores \gls{rigl}-level performance.}\label{fig:ablationvsnoablation}
\end{figure}

In brief, the methodology of \gls{rigl} is to update the \gls{snn} connectivity during training by \textit{pruning} weights with the smallest magnitude and \textit{regrowing} those with the largest corresponding gradient magnitude in \emph{each layer}. This occurs in periodic, but relatively infrequent mask update steps throughout most of training. 
In \gls{srigl}, weight saliency must be determined at the \emph{neuron level} (in convolutional layers, at the level of each filter), since we enforce that every neuron (output channel) has the same number of unmasked incoming weights, thereby satisfying the constant fan-in constraint. (\cref{fig:constfanin}). 

However, this approach alone significantly lags behind \gls{rigl}'s generalization at very high sparsities (>90\%) and with transformer architectures, as shown in \cref{fig:resnet50_acc_vs_sparsity,table:vit_imagenet_table}. This is because the constant fan-in constraint has an important side-effect: under a strict constant fan-in constraint, neurons can never be entirely masked (ablated), as illustrated in \cref{fig:ablationvsnoablation}. At very high sparsity levels this can lead to many neurons that have only 1--2 weights, limiting the capacity to learn complex features and consequently reducing generalization performance. Indeed, at high sparsities we observed empirically that \gls{rigl} ablates large numbers of neurons (\cref{fig:imagenet_perc_active,fig:resnet18_width,fig:vit-rigl-fan-in}). Effectively, \emph{\gls{rigl} reduces the width of the model at high sparsities to maintain generalization performance}; we believe we are the first to explicitly identify this behaviour within a \gls{dst} method.
To resolve this issue in \gls{srigl}, we implement a \textbf{neuron ablation method}, allowing \gls{srigl} to maintain both a constant fan-in constraint \emph{and} to reduce layer width at high sparsities. We introduce a new hyperparameter, $\gamma_{sal}$, which defines the required minimum percentage of salient weights per neuron. Given a neuron with constant fan-in of $k$, if fewer than $\gamma_{sal} * k$ weights are considered salient by either the drop \emph{or} grow criteria, then the neuron is ablated and its weights redistributed to other neurons within the same layer. 
Notably this neuron ablation method allows \gls{srigl} to exploit neuron ablation structured sparsity \emph{at much lower sparsity levels} than we identified it occurring at in \gls{rigl}, while maintaining good generalization, as demonstrated in \cref{table:vit_imagenet_table}.

The steps below outline our final \gls{srigl} method with neuron ablation. In the following procedure, the first two steps are the same as in \gls{rigl}, while the other steps are specific to \gls{srigl}, containing modifications to include the constant fan-in constraint and dynamic neuron ablation.
We first set an ablation threshold $\gamma_{sal}$. Then, for each layer we do the following:
\begin{enumerate}
    \item Obtain magnitudes of the active weights and gradient magnitudes of the pruned weights; these will serve as prune and growth criteria, respectively.
    \item Compute $K$, the number of weights to be grown and pruned in the current step in this layer. We always grow the same number of connections as we prune.
    \item Count the number of salient weights per neuron. A weight is considered \textit{salient} if it is in the top-$K$ of \emph{either} the largest-magnitude weights or the largest-magnitude gradients.
    \item Ablate neurons that have fewer salient weights than $\gamma_{sal} * k$, where $k$ is the fan-in. Ablation is done by pruning all incoming weights. These pruned weights are redistributed to the remaining neurons in the following steps.
    \item Compute the new constant fan-in constraint, $k^{\prime}$, based on the number of ablated neurons.
    \item Prune the $K$ smallest-magnitude weights in the current layer. Note that this pruning criterion considers all weights within a layer rather than pruning only the smallest weights in each neuron.
    \item For each active neuron, regrow as many weights as required, proceeding in order of decreasing gradient magnitude, until the target fan-in, $k^{\prime}$, is achieved.
\end{enumerate}
\section{Results}\label{sec:results}
We implement \gls{srigl} in PyTorch by extending an existing implementation of \gls{rigl}~\citep{nollied}.
We evaluate our method empirically on image classification tasks: on the CIFAR-10 dataset~\citep{krizhevsky_learning_2009} we train a variant of ResNet-18~\citep{he_deep_2016} suitable for CIFAR-10 and Wide ResNet-22~\citep{zagoruyko_wide_2017}; on the \gls{ilsvrc} dataset~\citep{imagenet} --- commonly referred to as ImageNet --- we train ResNet-50~\citep{he_deep_2016}, MobileNet-V3~\citep{howard_2019_iccv}, and \gls{vit}~\citep{dosovitskiy_image_2021}. See \cref{sec:wide_res_net} and \cref{sec:mobilenet} for Wide ResNet-22 and MobileNet-V3 experimental results, respectively.

Unless noted otherwise, we use the same hyperparameter configuration as the original \gls{rigl} method. A detailed summary of our hyperparameter settings and training details can be found in \cref{sec:training_details}. 

We set the ablation threshold, $\gamma_{sal}$, to 30\% for all \gls{srigl} results, except for our \gls{vit} experiments. This value was selected based on a hyperparameter sweep performed by training ResNet-18 and Wide ResNet-22 on the CIFAR-10 dataset, see \cref{sec:min_sal_weights_sweep}.

\subsection{ResNet-18 trained on CIFAR-10}\label{sec:resnet18}
We use a variant of ResNet-18 with reduced kernel dimensions and stride in the first two convolutional layers to obtain a model suitable for CIFAR-10; our training regimen generally follows \citet{evci_rigging_2021}, see \cref{sec:cifar_training_details} for more information. We repeat training with five different random seeds for both methods and report the mean and 95\% confidence interval compared to a densely-connected benchmark model in~\Cref{table:resnet18_cifar10_table}. These results confirm that imposing a constant fan-in constraint during sparse training does not significantly degrade generalization performance of the \gls{snn} compared to the \gls{rigl} method. In \cref{fig:resnet18_width} we plot the number of neurons ablated at ablation thresholds of 0\%, 30\%, and 50\% to demonstrate how the $\gamma_{sal}$ hyperparameter can be used to guide the final model width during training.

\subsection{ResNet-50 trained on ImageNet}
\begin{figure*}[t]
    \centering
    \begin{subfigure}{0.49\linewidth}
    \includegraphics[width=\linewidth]{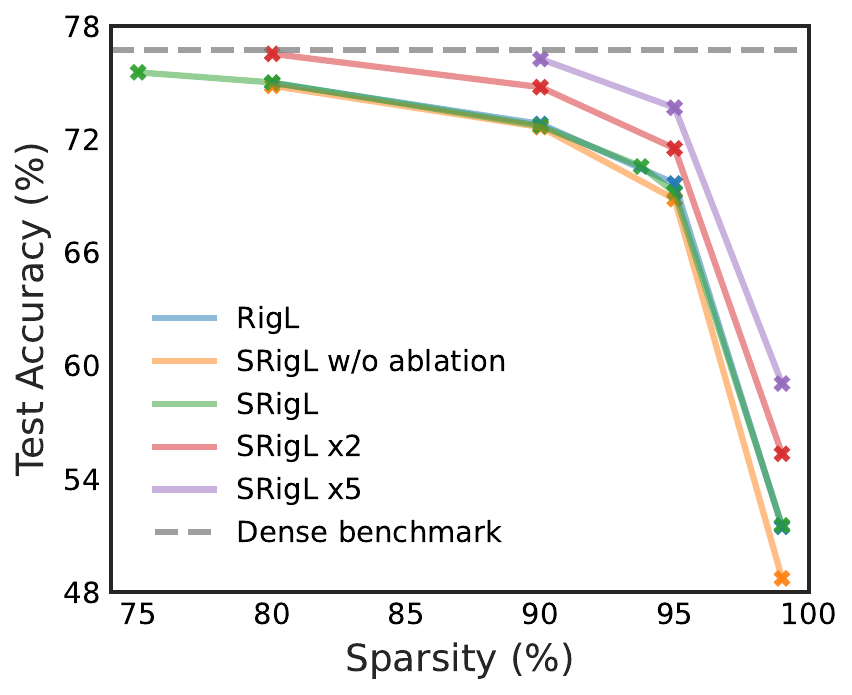}
    \caption{ResNet-50/ImageNet}\label{fig:resnet50_acc_vs_sparsity}
    \end{subfigure}
    \hfill
    \begin{subfigure}{0.49\linewidth}
    \includegraphics[width=\linewidth]{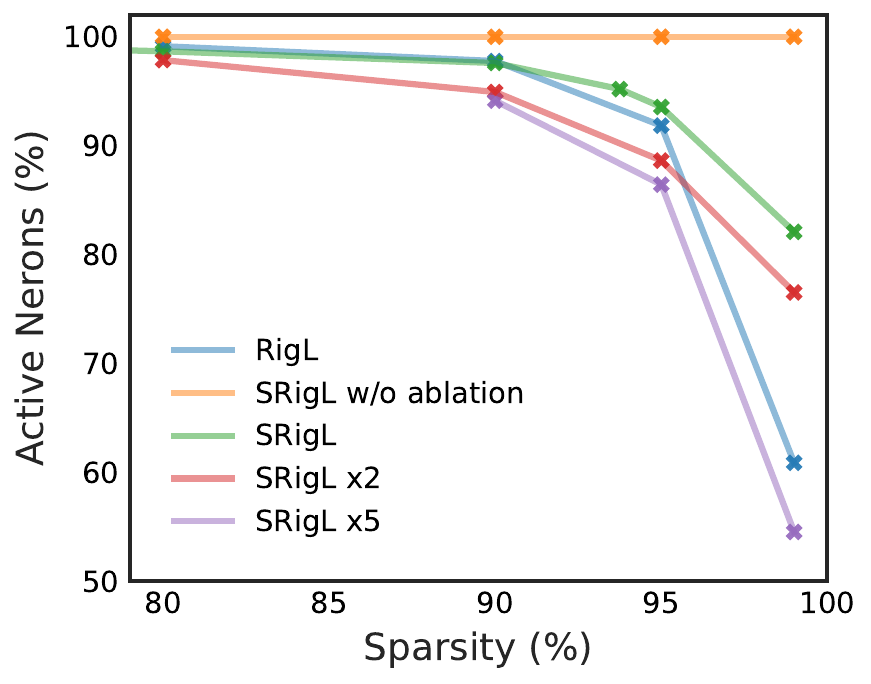}
    \caption{Neuron Ablation}\label{fig:imagenet_perc_active}
    \end{subfigure}
    \caption{\textbf{(\subref{fig:resnet50_acc_vs_sparsity}) \textbf{ResNet-50/ImageNet} top-1 test accuracy} when trained with \Gls{srigl} for a range of sparsities is comparable to \gls{rigl}. Extended training durations of $\times 2$ and $\times 5$ are also reported for \gls{srigl}. Results reported are single runs.
    \textbf{(\subref{fig:imagenet_perc_active}) Neuron ablation:} The percentage active neurons (i.e., not ablated) following \gls{rigl}/\gls{srigl} training on ResNet-50/ImageNet. \Gls{rigl} ablates a large number of neurons at high sparsities.}\label{fig:imagenet_resnet50_results}
\end{figure*}
Our training regimen for the ImageNet dataset generally follows~\citet{evci_rigging_2021}, see \cref{sec:resnet50_training_details} for more details. We investigate the effect of extended training with $\times$2 and $\times$5 the original number of training epochs. We train each model with a single seed and report the results in \cref{fig:resnet50_acc_vs_sparsity} and \cref{table:resnet50_imagenet_table}. 

\gls{srigl} yields similar generalization performance as \gls{rigl} across each sparsity and training duration considered. At high sparsities, \gls{srigl} with ablation outperforms \gls{srigl} without ablation, highlighting the importance of neuron ablation as sparsity increases. Notably, \gls{rigl}~$\times 5$ results at 99\% sparsity in \citet{evci_rigging_2021} used a dense first layer, unlike all other results reported in \cref{table:resnet50_imagenet_table}. Despite this difference, \gls{srigl}~$\times 5$ at 99\% sparsity is comparable to the \gls{rigl}~$\times 5$ results. We expect that the 99\% sparse models would be improved by using a dense first layer for all \gls{srigl} results. Similar to \gls{rigl}, we observe that \gls{srigl} generalization performance improves with increasing training time.

We inspect the connectivity of ResNet models trained with the \gls{rigl} method and find, as shown in \cref{fig:imagenet_perc_active}, that at 95\% sparsity $10.9\%$ of neurons are removed completely. Thus, \gls{rigl} results in fewer, but more densely connected neurons, whereas the fan-in constraint enforces that all neurons are retained.

In \cref{table:imagenet_benchmarks} we compare \gls{srigl} to a variety of \gls{dst} algorithms. \gls{srigl} performs comparably to other methods, even those which learn unstructured sparsity. Methods with a memory footprint listed as dense require training with the dense network and therefore are not directly comparable to other sparse-to-sparse \gls{dst} methods. The most directly comparable method to ours is \gls{dsb}; we note that \gls{srigl} outperforms \gls{dsb} at all sparsity ratios reviewed.
\begin{table}[hp]
\begin{minipage}[t]{.47\linewidth}
\renewcommand{\arraystretch}{0.45}
\begin{threeparttable}
\begin{center}
\caption{\textbf{Top-1 ImageNet test accuracy of ResNet-50} trained with \gls{rigl} or \gls{srigl} at high sparsities and with various training times (as in \citet{evci_rigging_2021}), e.g.\ 5$\times$ more training epochs than dense ResNet-50.}\label{table:resnet50_imagenet_table}
\begin{tabular}{@{}p{2em}cccccc@{}}
\toprule
 & \multicolumn{2}{c}{\gls{rigl}} & \multicolumn{4}{c}{\gls{srigl}}\\
\cmidrule(lr){2-3} \cmidrule(lr){4-7}
sparsity &&& \multicolumn{1}{c}{w/o} & \multicolumn{3}{c}{w/ ablation}\\
\cmidrule(lr){4-4} \cmidrule(lr){5-7}
(\%)& \multicolumn{1}{c}{1$\times$} & \multicolumn{1}{c}{5$\times$\tnote{\textdagger}} & \multicolumn{1}{c}{1$\times$} & \multicolumn{1}{c}{1$\times$} & \multicolumn{1}{c}{2$\times$} & \multicolumn{1}{c}{5$\times$}\\
\midrule
80 & $74.9$ & $77.1$ & $74.8$ & $75.0$ & $76.5$ & 77.2 \\\addlinespace
90 & $72.8$ & $76.6$ & $72.6$ & $72.7$ & $74.7$ & $76.2$ \\\addlinespace
95 & $69.6$ & $74.6$ & $68.8$ & $69.1$ & $71.5$ & $73.6$ \\\addlinespace
99 & $51.4$ & 61.9\tnote{\textdaggerdbl} & $48.7$ & $51.5$ & $55.3$ & $59.0$\\\addlinespace
\cmidrule{1-7}
0 & \multicolumn{6}{c}{\emph{dense ResNet-50}:\quad $76.7$}\\
\bottomrule
\end{tabular}
\begin{tablenotes}
\footnotesize
\item[\textdagger]~$5\times$ \gls{rigl} results are from \citet{evci_rigging_2021}
\item[\textdaggerdbl]~uses a dense first layer, unlike other results
\end{tablenotes}
\end{center}
\end{threeparttable}
\end{minipage}
\hfill
\begin{minipage}[t]{.47\linewidth}
\renewcommand{\arraystretch}{0.45}
\begin{center}
\caption{\textbf{Test accuracy for ResNet-18 on CIFAR-10} trained with RigL or \gls{srigl} with/without neuron ablation at varying sparsities repeated with five different random seeds.}\label{table:resnet18_cifar10_table}
\vspace*{0.3mm}
\begin{tabular}{@{}p{2em}ccc@{}}
\toprule
 & \gls{rigl} & \multicolumn{2}{c}{\gls{srigl}}\\
\cmidrule(lr){2-2} \cmidrule(lr){3-4}
sparsity & &  \multicolumn{1}{c}{w/o} & \multicolumn{1}{c}{w/ ablation}\\
\cmidrule(lr){3-3} \cmidrule(lr){4-4}
(\%) & & & \\
\\[-0.33em]
\midrule
80 & $95.2\pm 0.1$ & $95.2\pm 0.1$ & $95.2\pm 0.0$ \\ \addlinespace
90 & $95.1\pm 0.1$ & $95.0\pm 0.1$ & $95.1\pm 0.1$ \\ \addlinespace
95 & $94.6\pm 0.2$ & $94.5\pm 0.3$ & $94.7\pm 0.2$ \\ \addlinespace
99 & $92.9\pm 0.1$ & $91.5\pm 0.3$ & $92.8\pm 0.1$ \\ \addlinespace
\\[-0.55em]
\cmidrule{1-4} 
0 & \multicolumn{3}{c}{\emph{dense ResNet-18}:\quad $95.5$}\\
\bottomrule
\end{tabular} 
\end{center}
\end{minipage}
\end{table}
\begin{table}[htbp]
\begin{threeparttable}
\begin{center}
    \caption{\textbf{Top-1 ImageNet test accuracy of ResNet-50 trained with a variety of \gls{dst} methods}, highlighting methods that both are sparse-to-sparse (i.e.\ sparse training) \emph{and} learn structured sparsity similar to \gls{srigl} --- only DSB-16 (2:4 and 1:4  sparsity) is directly comparable in this regard. \gls{rigl} and \gls{srigl} results are from our experiments, other values are obtained from each method's corresponding paper, unless noted otherwise. }\label{table:imagenet_benchmarks}
    \begin{tabular}{@{}p{5.5em}p{3.5em}p{4em}cccccc@{}}\toprule
    & training &  & \multicolumn{5}{c}{sparsity} \\
    \cmidrule(lr){4-8}
    method & method & structured & 50\% & 75\% & 80\% & 90\%  & 93.75\% \\
    \midrule
    Static\tnote{\textasteriskcentered} & sparse & no & -- & -- & $70.6\pm 0.06$ & $65.8\pm 0.04$ & -- \\ 
    SET\tnote{\textasteriskcentered} & sparse & no & -- & -- & $72.9\pm 0.39$ & $69.6\pm 0.23$ & -- \\  
    DeepR\tnote{\S} & sparse & no & -- & -- & $71.7$ & $70.2$ & -- \\ 
    DSR & sparse & no & -- & -- & $73.3$ & $71.6$ & -- \\  
    Top-KAST\tnote{\textdaggerdbl} & sparse & no & -- & -- & $74.76$ & $70.42$ & -- \\ 
    MEST\tnote{\textdagger} & sparse & no & -- & -- & $\mathbf{75.39}$ & $72.58$ & -- \\ 
    RigL & sparse & no & -- & -- & $74.98$ & $72.81$ & -- \\ 
    DSB-16 & \textbf{sparse} & \textbf{yes} & $76.33$ & $74.04$ & -- & -- & -- \\ 
    Chase \tnote{\textdagger\textdagger} & \textbf{sparse} & \textbf{yes} & -- & -- & $75.27$ & $\mathbf{74.03}$ & -- \\
    SRigL (Ours) & \textbf{sparse} & \textbf{yes} & $\mathbf{76.60}$ & $\mathbf{75.55}$ & $75.01$ & $72.71$ & $70.56$ \\ 
    \cmidrule{1-8}
    SNFS (ERK)\tnote{\textasteriskcentered} & dense & no & -- & -- & $75.2\pm 0.11$ & $73.0\pm 0.04$ & -- \\ 
    AST+GC\tnote{\textasteriskcentered\textasteriskcentered} & dense & no &  --  & -- & $73.2$ & $73.1$ & -- \\
    SR-STE & dense & yes & -- & $76.2$ & -- & -- & $71.5$ \\ 
    \cmidrule{1-8}
    \multicolumn{3}{c}{\emph{dense ResNet-50:}} & \multicolumn{5}{c}{$76.7$} \\\bottomrule
\end{tabular}
\begin{tablenotes}
\footnotesize
\item \textasteriskcentered Values obtained from \citet{evci_rigging_2021}. \S values obtained from \citet{mostafa_parameter_2019}. \textdagger Values for the MEST (x0.67+EM) variant, matched to the same number of training \gls{flops} as \gls{rigl}. \textdaggerdbl Values tabulated for \gls{topkast} correspond to the \emph{backwards sparsity} as \gls{topkast} uses different sparsities in the forward and backward passes. For more information see Table 1 in \citet{jayakumar_top-kast_2020}. \textdagger\textdagger Values from \citet{yin_dynamic_2023} for channel sparsity ($S_c$) set to 40\%. \textasteriskcentered\textasteriskcentered 50\% initial sparsity. Values from \citet{yang_get_2022}
\end{tablenotes}
\end{center}
\end{threeparttable}
\end{table}

\begin{table}[h!bp]
\begin{minipage}[t]{0.45\linewidth}
\renewcommand{\arraystretch}{0.45}
\begin{threeparttable}
\begin{center}
\caption{{\bf Top-1 test accuracy of ViT-B/16} trained on ImageNet with or w/o neuron ablation}\label{table:vit_imagenet_table}
\begin{tabular}{@{}cccc@{}}
\toprule
& \gls{rigl} & \multicolumn{2}{c}{\gls{srigl}} \\ 
\cmidrule(lr){2-2} \cmidrule(lr){3-4}
sparsity (\%)\tnote{\textdagger} &  & w/o & w/ ablation \\ 
\midrule
80 & $\mathbf{77.9}$ & $73.5$ & $77.5$ \\ \addlinespace
90 & $\mathbf{76.4}$ & $71.3$ & $76.0$ \\ \addlinespace
\midrule 
0 & \multicolumn{3}{c}{\emph{dense \gls{vit}}: $78.35$}\\ 
\bottomrule 
\end{tabular} 
\begin{tablenotes}
\footnotesize
    \item \textdagger Sparsity level set for all modules \emph{except multi-headed attention input projections}, which remain dense. See \cref{sec:vit_training_details} for more details. 
\end{tablenotes}
\end{center}
\end{threeparttable}
\end{minipage}
\hspace{0.2em}
\begin{minipage}[t]{0.45\linewidth}
\renewcommand{\arraystretch}{0.45}
\begin{center}
\caption{\textbf{\gls{srigl} sparsity and \gls{flops}} for ResNet-50/ImageNet training  and inference. See \cref{sec:flops} for more details.}\label{table:flops}
\begin{tabular}{@{}ccc@{}}
\toprule
& \multicolumn{2}{c}{\gls{srigl} \gls{flops}}\\
\cmidrule(lr){2-3}
sparsity (\%) &  training ($\times 1e18$) & inference ($\times 1e9$) \\
\midrule
80 & $1.13$ &  $3.40$ \\ \addlinespace
90 & $0.77$ & $1.99$ \\ \addlinespace
95 & $0.40$ & $1.01$  \\ \addlinespace
99 & $0.09$ & $0.21$ \\ \addlinespace
\midrule 
0 & $3.15$ & $8.20$\\
\bottomrule
\end{tabular} 
\end{center}
\end{minipage}
\end{table}

\FloatBarrier

\subsection{Vision Transformer trained on ImageNet}\label{sec:vit}
We train the vision transformer variant \gls{vit} on ImageNet generally following the original training recipe per \citet{dosovitskiy_image_2021} with select modifications, see \cref{sec:vit_training_details} for more information.

Similar to our \gls{cnn} experiments, \gls{rigl} ablates a significant number of neurons when applied to the \gls{vit} architecture with sparsities of  80 and 90\%. Additionally, we find that \gls{rigl} learns sparse connectivities with a high variance of fan-in between neurons (see \cref{fig:vit-rigl-fan-in}). At 90\% sparsity, some neurons are allocated up to $\times 10$ more active weights than the mean number of active weights in the same layer. We hypothesize that these more densely connected neurons found in our \gls{rigl} experiments are important for generalization performance; therefore, a high  $\gamma_{sal}$ threshold should improve performance of \gls{srigl} by ablating neurons until a sufficient density of sparse fan-in is reached. Indeed, we find that \gls{srigl}'s generalization performance is sensitive to $\gamma_{sal}$ and that high $\gamma_{sal}$ thresholds of 90\% to 99\% perform best. See \cref{fig:vit_min_sal,sec:min_sal_weights_sweep} for more details on how $\gamma_{sal}$ affects the generalization performance of \gls{vit}. For the following results, we used a $\gamma_{sal}$ of 95\%.

We train each model with a single random initialization and report the results in \cref{table:vit_imagenet_table}. \gls{srigl} without ablation is unable to match the generalization performance of \gls{rigl} at very high sparsity. However, with neuron ablation enabled, \gls{srigl}'s performance greatly improves and is closely comparable to \gls{rigl} at 80\% and 90\% sparsity.
\subsection{Acceleration of constant fan-in sparsity}\label{sec:accel}
\newcommand{\pluseq}{\mathrel{{+}{=}}}
\newcommand{\code}[1]{\texttt{#1}}
\begin{algorithm}
    \small
    \begin{algorithmic}[1]
    \caption{``Condensed'' linear layer with constant fan-in sparsity forward pass}\label{alg:unrolled}
        \State \textbf{Input:} \texttt{x}: the input matrix of shape (\texttt{batch\_size}, \texttt{num\_features)} \\
            \qquad \quad \texttt{w}: the condensed weight matrix of shape (\texttt{active\_neurons}, \texttt{constant\_fan\_in)} \\
            \qquad \quad \texttt{indx}: indices of non-zero dense weights of shape (\texttt{active\_neurons},\\ \qquad \quad\texttt{constant\_fan\_in})
        \State \texttt{output} $\gets$ \texttt{torch.zeros(size=(batch\_size,\ neurons))}
        \For{\texttt{b in range(batch\_size)}} \Comment{For each sample in mini-batch}
            \For{\texttt{n in range(neurons)}} \Comment{For each active neuron in layer}
                \For{\texttt{k in range(constant\_fan\_in)}} \Comment{For each non-zero weight}
                    \State \texttt{source\_idx $\gets$ idx[n, k]}
                    \State \texttt{feature $\gets$ x[b, source\_idx]}
                    \State \texttt{output[b,\ n] $\pluseq$ feature $*$ w[n,\ k]}
                \EndFor
            \EndFor
        \EndFor
    \State \textbf{return} \texttt{output}
    \end{algorithmic}
\end{algorithm}
\begin{figure*}[tbp]
    \centering
\begin{subfigure}{0.95\linewidth}
    \includegraphics[width=\linewidth]{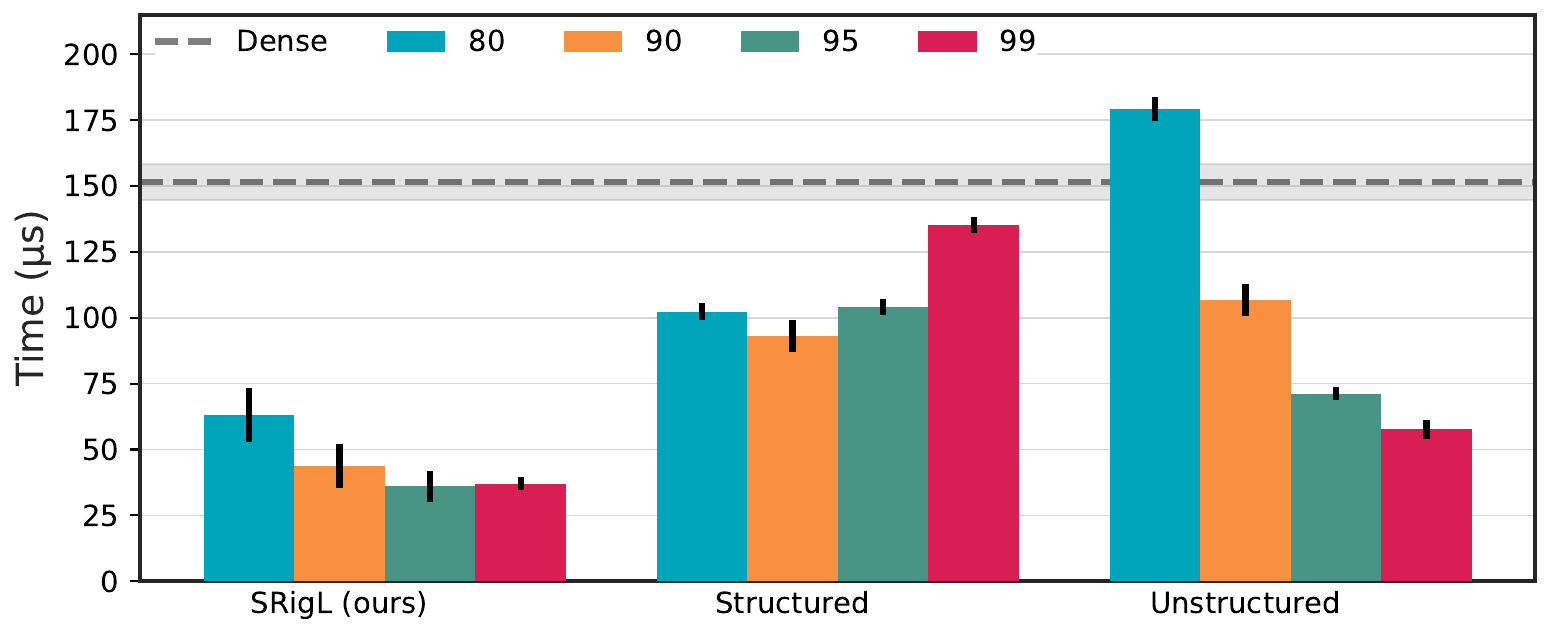}
    \caption{\gls{cpu} online inference}\label{fig:online-inference-timings}
\end{subfigure}
\begin{subfigure}{0.95\linewidth}
        \includegraphics[width=\linewidth]{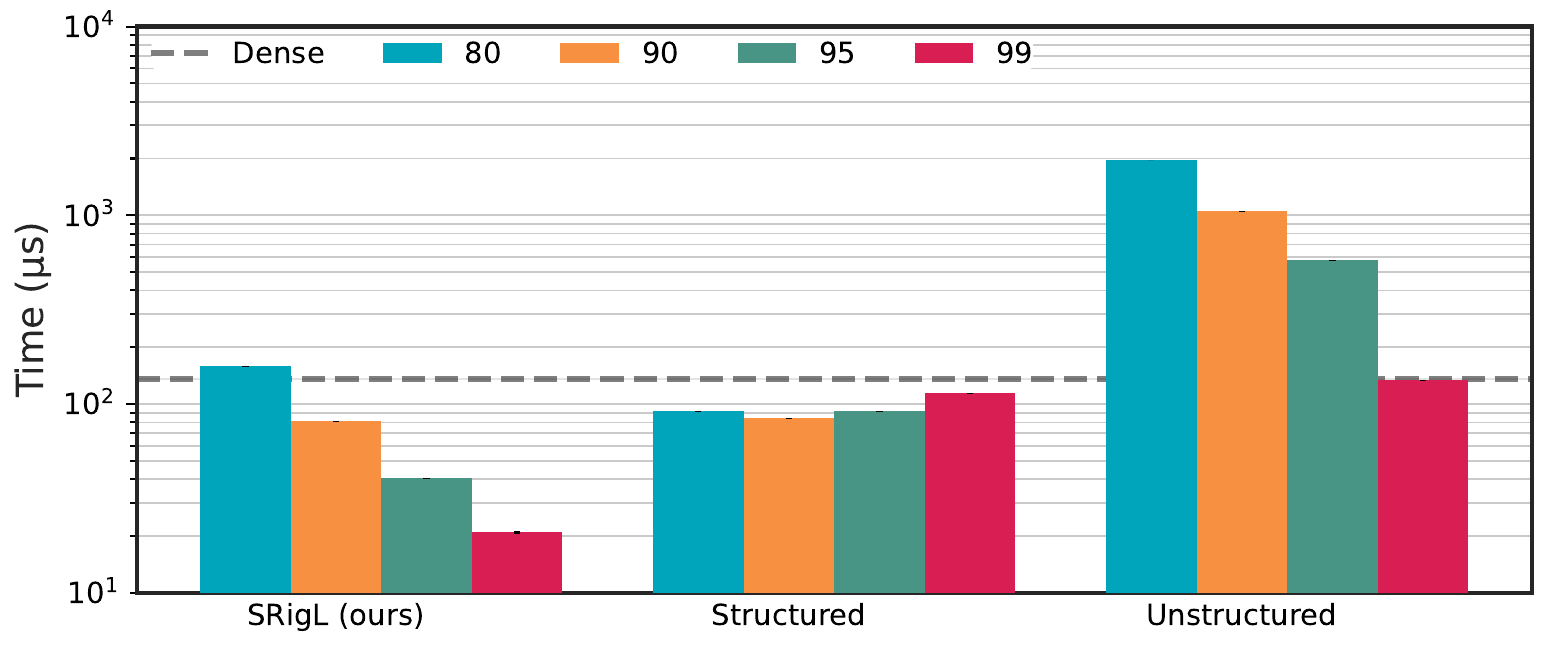}
        \caption{GPU inference with batch size of 256}\label{fig:gpu-accel-bs-256}
\end{subfigure}
    \caption{\textbf{Comparing real-world timings for a fully-connected layer} extracted from a \gls{vit} model trained with \gls{srigl} when compressed using the condensed representation learned by \gls{srigl} to structured (i.e.\ the same layer accelerated using only the ablated neurons without exploiting the fine-grained sparsity), and unstructured (i.e.\ \gls{csr}) representations. The median over a minimum of 5 runs is shown, while the error bars show the std.\ dev. \emph{\textbf{Note:} the increased timings for the 95 \& 99\% sparse structured representations is due to \gls{srigl} ablating relatively fewer neurons at these sparsities compared to 80 and 90\%.}
    \textbf{(\subref{fig:online-inference-timings}) CPU wall-clock timings for online inference} on an Intel Xeon W-2145. For online (single input) inference, our condensed representation at 90\% is \emph{3.4$\times$ faster than dense} and \emph{2.5 $\times$ faster than unstructured sparsity}. See \cref{sec:timingsdetails}.
    \textbf{(\subref{fig:gpu-accel-bs-256}) GPU wall-clock timings for inference with a batch size of 256} on an NVIDIA Titan V. At 90\% sparsity, our condensed representation is \emph{1.7$\times$ faster than dense} and \emph{13.0$\times$ faster than unstructured (\gls{csr})} sparse layers. Note y-axis is log-scaled.} \label{fig:timings}
\end{figure*}
While \gls{srigl} shows promising theoretical speedups (i.e.\ \gls{flops}) as demonstrated in \cref{table:flops,sec:flops}, \gls{flops} are limited in demonstrating the real-world acceleration potential of a proposed sparse representation in general. Yet conversely, creating a fully-optimized software or hardware implementation of a novel representation typically requires significant engineering effort outside of the scope of this paper.

Here we show that even a straight-forward PyTorch implementation of our proposed condensed neural network representation (see \cref{sec:condensedrepresentationmath}) can demonstrate this real-world acceleration. The algorithm to accelerate our condensed sparsity representation is shown in \cref{alg:unrolled}, demonstrating that it is embarrassingly parallel. Additionally, leveraging CUDA kernels from \citet{schultheis_towards_2023}, we also demonstrate that constant fan-in sparsity can be accelerated on commodity \glspl{gpu}. 

To accelerate our condensed linear layer we exploit both structured and constant fan-in sparsity by removing ablated neurons and zero-valued weights from active neurons. In \cref{fig:timings}, we present real-world timings comparing our condensed linear layer to structured and unstructured sparse representations. 
We extract the trained layer weights and bias from \gls{vit} models trained with \gls{srigl} to obtain an accurate representation of the sparse topology produced during a real training run with \gls{srigl}.

Our condensed representation is significantly faster than the dense benchmark and other sparse representations across all sparsities investigated. This real-world speed-up is immediately applicable to applications where latency is critical. In some instances, we found structured sparsity yields the best acceleration. By including both structured and constant fan-in sparsity, models trained with \gls{srigl} can use either the fully condensed (structured + constant fan-in) \emph{or} purely structured sparse representations to obtain real-world acceleration across a broad range of applications with the \emph{same set of weights}. See \cref{sec:timingsdetails} and \cref{sec:gpu-benchmarks} for details on wall-clock benchmarks across a range of threads and batch sizes. Furthermore, we expect that a more optimized software implementation and/or explicit hardware support would enable use of \gls{srigl} across a wider range of applications.
\section{Conclusion}
In this work we present \gls{srigl}, a novel \gls{dst} method that learns a sparsity mask incorporating both structured and constant fan-in sparsity. \gls{srigl} is capable of sparse-to-sparse training while maintaining generalization performance on par with state-of-the-art unstructured sparse training methods on a wide variety of network architectures. Our observation that \gls{rigl} ablates neurons at high sparsities inspires our neuron ablation method which enables \gls{srigl} to match the performance of \gls{rigl}, even at high sparsities and on the \gls{vit} network architecture. \gls{srigl}'s constant fan-in constraint and neuron ablation results in real-world acceleration for \gls{cpu} online inference and \gls{gpu} batched inference. We hope this work will motivate the implementation of additional fine-grained structured sparsity schemes and the engineering efforts required to accelerate them further. 

\subsubsection*{Acknowledgments}
We acknowledge the support of  Alberta Innovates, the Natural Sciences and Engineering Research Council of Canada (NSERC), and the NSF AI Institute for Artificial Intelligence and Fundamental Interactions (IAIFI). We are grateful for computational resources made available to us by Denvr Dataworks, Google, Amazon, and the Digital Research Alliance of Canada. We also acknowledge the very helpful feedback of Erik Schultheis and Trevor Gale. 

\bibliography{mybib,chase-bib}
\bibliographystyle{iclr2024_conference}
\newpage
\appendix
\appendices
\addappheadtotoc
\appendixpage
\section{Sparsity and Output-Norm Variance}\label{sec:outputnormvariance}
Consider a \gls{snn} with ReLU activations, where each neuron has on average $k$ connections to the previous layer (i.e., fan-in). 
It has been shown by~\citet{yani2022}, that by normalizing the weights on initialization by a factor of $\sqrt{2/k}$, 
one achieves the following desirable normalization property for each layer $\ell$ with output $z^\ell$:
$$\mathbb{E}\left(\frac{||z^{\ell+1}||^2}{||z^{\ell}||^2} \right)= 1\,,$$  
Meaning that on average the variance of the norm of each layer's output is constant. However, the variance of this ratio is non-trivial. 
In networks with large depth, it can accumulate, leading to exponentially large variance at the final layer~\citep{mihai2021}. 
Minimizing this variance on initialization has been shown to have a positive effect on training dynamics in some network models~\citep{collegialnetworks}, as it stabilizes the gradients. We therefore analyze the output norm variance as a guiding quantity for sparsity-type selection. 

In the following, we consider three different types of sparsity distributions, which respectively correspond to different degrees of sparsity \textit{structure} in the \gls{snn}, and derive analytic expressions for the behaviour of output norm variance in \glspl{snn} with the given sparsity type. The derivations for the following results can be found in~\cref{appx:math}:

\begin{itemize}[itemsep=0.2em, leftmargin=2em]
\item \textbf{``Bernoulli sparsity''}: A connection between each neuron in layer $\ell+1$ and each neuron in layer $\ell$ appears \textit{independently} with probability $p=\frac{k}{n}$, resulting in each neuron having $k$ connections \textit{on average} and each layer having $nk$ connections \textit{on average}. The variance is:
\begin{equation}
    \mathbf{Var}_\textrm{Bernoulli}\left(\frac{||z^{\ell+1}||^2}{||z^{\ell}||^2} \right) = \frac{5n - 8 + 18 \frac{k}{n} }{n(n+2)}\,.
\end{equation}
\item \textbf{``Constant Per-Layer sparsity''}: Exactly $kn$ connections are distributed at random in the layer connecting the $n$ neurons in layer $\ell+1$ and the $n$ neurons in layer $\ell$, resulting in each neuron having $k$ connections \textit{on average}. The variance is:
\begin{eqnarray}
    \mathbf{Var}_\textrm{Const-Per-Layer}\left(\frac{||z^{\ell+1}||^2}{||z^{\ell}||^2} \right) =
    \frac{(n^2 + 7n -8)C_{n,k} + 18\frac{k}{n} - n^2 - 2n}{n(n+2)}\,,
\end{eqnarray}
where $C_{n,k} = \frac{n - 1/k}{n - 1/n}$. Note that when $n \gg 1$, $C_{n,k} \approx 1 - \frac{n-k}{n^2 k}$ is close to $1$, and with $C_{n,k}=1$ we recover the formula for Bernoulli sparsity, meaning that this sparsity type and Bernoulli sparsity are very similar.
\item \textbf{``Constant Fan-In sparsity''}: Each neuron in layer $\ell+1$ is connected to exactly $k$ neurons from layer $\ell$, chosen uniformly at random. In this case, the variance is:
\begin{eqnarray}
    \mathbf{Var}_\textrm{Const-Fan-In}\left(\frac{||z^{\ell+1}||^2}{||z^{\ell}||^2} \right) =
    \frac{5n - 8 + 18 \frac{k}{n} }{n(n+2)} - \frac{3(n-k)}{kn(n+2)}\,.
\end{eqnarray}
\end{itemize}
In deriving the above results we assumed that the direction of the layer output vector $\frac{z^\ell}{||z^\ell||}$ is uniformly distributed on the unit sphere. We compare our theoretical predictions with simulations in \cref{fig:theory} and verify their accuracy. 
Bernoulli and constant-per-layer distribution result in unstructured sparsity, and most of the current \gls{dst} approaches, including \gls{rigl}, operate with constant-per-layer sparsity. In contrast, the constant-fan-in type imposes a strong structural constraint. Therefore we are somewhat surprised to find that, in fact, constant-fan-in sparsity always produces slightly smaller output-norm variance than the other types. The difference is larger when $k \ll n$, i.e., for very sparse networks. This indicates that, at the very least, the constant fan-in constraint should not impair \gls{snn} training dynamics and performance, motivating our method of maintaining the constant fan-in sparsity constraint within a \gls{dst} approach.

\newpage
\section{Computing the output norm variance}\label{appx:math}
\newcommand{\var}{\mathbf{Var}}
\newcommand{\cov}{\mathbf{Cov}}
\newcommand{\dequal}{\overset{d}{=}}
\newcommand{\norm}[1]{\left\lVert#1\right\rVert}

\begin{definition}
Let $\xi\in\{0,1\}^{N}$ be a binary vector. Let $I\in\{0,1\}^{N\times N}$ be an $N\times N$ binary matrix. Let $u\in\mathbb{R}^{N}$ be any vector. Let $W\in\mathbb{R}^{N\times N}$ be a matrix of iid $\mathcal{N}(0,1)$ random variables.

Define the vector $z$ by:
\begin{equation}
    z=\sqrt{\frac{2}{k}}\left(W\odot I\right)\left(\xi\odot u\right)
\end{equation}
i.e. the entries $z_{i}$ are given by:
\begin{equation}
    z_{i}=\sqrt{\frac{2}{k}}\sum_{j=1}^{n}W_{ij}I_{ij}\xi_{j}u_{j}
\end{equation}
\end{definition}
\begin{proposition}
    The variance of each entry $z_{i}$ is:
    \begin{equation}
    \var(z_{i})	=\frac{2}{k}\sum_{j=1}^{n}I_{ij}\xi_{j}u_{j}^{2}
    \end{equation}
    and therefore the distribution of each $z_{i}$ can be written as 
    \begin{equation}
    z_{i}\dequal g_{i}\sqrt{\frac{2}{k}\sum_{j=1}^{n}I_{ij}\xi_{j}u_{j}^2}
    \end{equation}
    where $g_{i}$ are $N$ iid $\mathcal{N}(0,1)$ random variables.
\end{proposition}
\begin{proof}
By the properties of variance:
\begin{align}
    \var(z_{i})	&=\frac{2}{k} \sum_{j,j^{\prime}} I_{ij}I_{ij^{\prime}} \xi_{j}\xi_{j^{\prime}} u_{j}u_{j}^{\prime} \cov(W_{ij},W_{ij^{\prime}})\\
	&=\frac{2}{k} \sum_{j,j^{\prime}} I_{ij}I_{ij^{\prime}} \xi_{j}\xi_{j^{\prime}} u_{j}u_{j}^{\prime}\delta_{j=j^{\prime}}\\
	&=\frac{2}{k}\sum_{j} I_{ij}^{2} \xi_{j}^{2} u_{j}^{2}\\
	&=\frac{2}{k}\sum_{j} I_{ij} \xi_{j} u_{j}^{2}
\end{align}
since $I_{ij}^{2}=I_{ij}$ and $\xi_{j}^{2}=\xi_{j}$ because they are binary valued. Once the variance is established, notice that $z_{i}$ is a linear combination of Gaussians with $z_{i}\perp z_{i^{\prime}}$, because the row $W_{ij}\perp W_{i^{\prime}j}$. Hence the $z_{i}$ are independent Gaussians, so the form $z_{i}\dequal g_{i}\sqrt{\frac{2}{k}\sum_{j=1}^{n}I_{ij}\xi_{j}u_{j}^{2}}$ follows.
\end{proof}
\begin{corollary}
    The norm $\norm{z}^2$ can be written as:
    \begin{equation}
        \norm{z}^2\dequal\frac{2}{k}\sum_{i,j=1}^{n}g_{i}^{2}I_{ij}\xi_{j}u_{j}^{2}
    \end{equation}
\end{corollary}
\begin{proposition}[``Bernoulli Sparsity''] Suppose that $u\in\mathbb{R}^{n}$ is uniform from the unit sphere, the entries $I_{ij}\sim Ber\left(\frac{k}{n}\right)$, $\xi_{j}\sim Ber(\frac{1}2)$ all independent of each other. Then:
\begin{align}
    \mathbb{E}\left(\norm{z}^2\right) &= 1 \\
    \var\left(\norm{z}^2\right)	&= \frac{5n-8+18\frac{n}{k}}{n(n+2)}
\end{align}
\end{proposition}
\begin{proof}
    We have
    \begin{align}
    \mathbb{E}\left(\norm{z}^2\right)	
        &=\frac{2}{k}\sum_{i,j=1}^{n}\mathbb{E}\left[g_{i}^{2}I_{ij}\xi_{j}u_{j}^{2}\right]\\
	    &=\frac{2}{k}\sum_{i,j=1}^{n} \mathbb{E}\left[g_{i}^{2}\right]
                                      \mathbb{E}\left[I_{ij}\right]
                                      \mathbb{E}\left[\xi_{j}\right]
                                      \mathbb{E}\left[u_{j}^{2}\right]\\
	   &=\frac{2}{k}\sum_{i,j=1}^{n}1\cdot\frac{k}{n}\cdot\frac{1}{2}\cdot\frac{1}{n}\\
	   &=1
    \end{align}
Similarly, we compute the 4-th moment as follows:
\begin{equation}
    \mathbb{E}\left(\norm{z}^4\right) = \left(\frac{2}{k}\right)^{2}\sum_{i,j,i^{\prime},j^{\prime}}^{n}\mathbb{E}\left[g_{i}^{2}g_{i^{\prime}}^{2}\right]\mathbb{E}\left[I_{i^{\prime}j^{\prime}}I_{ij}\right]\mathbb{E}\left[\xi_{j}\xi_{j^{\prime}}\right]\mathbb{E}\left[u_{j}^{2}u_{j^{\prime}}^{2}\right]
\end{equation}
We split this into four cases and evaluate these based on whether or not $i=i^{\prime}$ and $j=j^{\prime}$ in the following table.
\begin{table}[tbp]
\centering
\begin{tabular}{cccccc}
\toprule
Case & Num. Terms & $\mathbb{E}\left[g_{i}^{2}g_{i^{\prime}}^{2}\right]$ & $\mathbb{E}\left[I_{i^{\prime}j^{\prime}}I_{ij}\right]$ & $\mathbb{E}\left[\xi_{j}\xi_{j^{\prime}}\right]$ & $\mathbb{E}\left[u_{j}^{2}u_{j^{\prime}}^{2}\right]$ \\
\midrule
$i=i^{\prime},j=j^{\prime}$ & $n^{2}$ & 3 & $\frac{k}{n}$ & $\frac{1}{2}$ & $\frac{3}{n(n+2)}$ \\
$i\neq i^{\prime},j=j^{\prime}$ & $n^{2}(n-1)$ & 1 & $\left(\frac{k}{n}\right)^{2}$ & $\frac{1}{2}$ & $\frac{3}{n(n+2)}$ \\
$i=i^{\prime},j\neq j^{\prime}$ & $n^{2}(n-1)$ & 3 & $\left(\frac{k}{n}\right)^{2}$ & $\left(\frac{1}{2}\right)^{2}$ & $\frac{1}{n(n+2)}$ \\
$i\neq i^{\prime},j\neq j^{\prime}$ & $n^{2}(n-1)^{2}$ & 1 & $\left(\frac{k}{n}\right)^{2}$ & $\left(\frac{1}{2}\right)^{2}$ & $\frac{1}{n(n+2)}$ \\
\bottomrule
\end{tabular}
\caption{Overview of terms for Bernoulli type sparsity.}
\label{tab:ber}
\end{table}

Combining the value of each term with the number of terms gives the desired result for the variance.
\end{proof}

\begin{proposition}[``Constant-per-layer sparsity''] Suppose that $u\in\mathbb{R}^{n}$ is uniform from the unit sphere and $\xi_{j}\sim Ber(\frac{1}2)$ are independent of each other. Suppose the entries of the matrix $I_{ij}$ are chosen such that:

There are exactly $kn$ ones and exactly $n^2-nk$ zeros in the matrix $I$, and their positions in the matrix are chosen uniformly from the $\binom{n^{2}}{nk}$ possible configurations. Then:
\begin{align}
    \mathbb{E}\left(\norm{z}^2\right) &= 1\\
    \var\left(\norm{z}^2\right)	&= \frac{(n^{2}+7n-8)C_{n,k} + 18\frac{k}{n} - n^{2}-2n}{n(n+2)}
\end{align}
\end{proposition}
\begin{proof}
    Note that $\mathbb{E}(I_{ij}) = k/n$ still holds, since there are $kn$ ones distributed over $n^2$ locations. Thus the computation for $\mathbb{E}(\norm{z}^2)$ is identical to the previous proposition. Note also that when there are two entries, we have:
    \begin{align}
        \mathbb{E}[I_{ij}I_{i^\prime j^\prime}]     = \begin{cases}
        \frac{k}n                                   &\text{if } i=i^\prime \text{ and } j=j^\prime \\
        \frac{k}n \cdot \frac{nk-1}{n^2-1}          &\text{otherwise}
        \end{cases}\\
        =
        \begin{cases}
        \frac{k}n                                   &\text{if } i=i^\prime \text{ and } j=j^\prime \\
        \left(\frac{k}n\right)^2 \cdot C_{n,k}      &\text{otherwise}
        \end{cases}
    \end{align}
where $C_{n,k} = \frac{n-1/k}{n-1/n}$. The table with terms for computing $\mathbb{E}(\norm{z}^4)$ becomes:
\begin{table}[tbp]
\centering
\begin{tabular}{cccccc}
\toprule
Case & Num. Terms & $\mathbb{E}\left[g_{i}^{2}g_{i^{\prime}}^{2}\right]$ & $\mathbb{E}\left[I_{i^{\prime}j^{\prime}}I_{ij}\right]$ & $\mathbb{E}\left[\xi_{j}\xi_{j^{\prime}}\right]$ & $\mathbb{E}\left[u_{j}^{2}u_{j^{\prime}}^{2}\right]$ \\
\midrule
$i=i^{\prime},j=j^{\prime}$ & $n^{2}$ & 3 & $\frac{k}{n}$ & $\frac{1}{2}$ & $\frac{3}{n(n+2)}$ \\
$i\neq i^{\prime},j=j^{\prime}$ & $n^{2}(n-1)$ & 1 & $\left(\frac{k}{n}\right)^{2}C_{n,k}$ & $\frac{1}{2}$ & $\frac{3}{n(n+2)}$ \\
$i=i^{\prime},j\neq j^{\prime}$ & $n^{2}(n-1)$ & 3 & $\left(\frac{k}{n}\right)^{2}C_{n,k}$ & $\left(\frac{1}{2}\right)^{2}$ & $\frac{1}{n(n+2)}$ \\
$i\neq i^{\prime},j\neq j^{\prime}$ & $n^{2}(n-1)^{2}$ & 1 & $\left(\frac{k}{n}\right)^{2}C_{n,k}$ & $\left(\frac{1}{2}\right)^{2}$ & $\frac{1}{n(n+2)}$ \\
\bottomrule
\end{tabular}
\caption{Overview of terms for Constant-per-layer type sparsity.}
\label{tab:tab2}
\end{table}
\vspace{1cm}
The extra factor of $C_{n,k}$ in the entries leads to the stated result.
\end{proof}

\begin{proposition}[``Constant Fan-In sparsity''] Suppose that $u\in\mathbb{R}^{n}$ is uniform from the unit sphere, and $\xi_{j}\sim Ber(\frac{1}2)$ all independent of each other. Suppose the entries of the matrix $I_{ij}$ are chosen so that:
\begin{enumerate}
    \item There are exactly $k$ ones \textbf{in each row} of the matrix $I$ and exactly $n-k$ zeros in the matrix $I$, chosen uniformly from the $\binom{n}{k}$ possible ways this can happen.
    \item Different rows of $I$ are independent.
\end{enumerate}
Then:
\begin{align}
    \mathbb{E}\left(\norm{z}^2\right) &= 1\\
    \var\left(\norm{z}^2\right)	&= \frac{5n-8+18\frac{n}{k}}{n(n+2)}-\frac{3(n-k)}{kn(n+2)}
\end{align}
\end{proposition}
\begin{proof}
    Same arguments as before apply, but now we have
    \begin{align}
        \mathbb{E}[I_{ij}I_{i^\prime j^\prime}] = \begin{cases}
        \frac{k}n                       &\text{if } i=i^\prime \text{ and } j=j^\prime \\
        \frac{k}n \frac{k-1}{n-1}       &\text{if } i=i^\prime \text{ and } j\neq j^\prime \\
        \left(\frac{n}{n}\right)^2      &\text{otherwise}
        \end{cases}\\
    \end{align}
and the table for the variance computation becomes:
\begin{table}[!htbp]
\centering
\begin{tabular}{cccccc}
\toprule
Case & Num. Terms & $\mathbb{E}\left[g_{i}^{2}g_{i^{\prime}}^{2}\right]$ & $\mathbb{E}\left[I_{i^{\prime}j^{\prime}}I_{ij}\right]$ & $\mathbb{E}\left[\xi_{j}\xi_{j^{\prime}}\right]$ & $\mathbb{E}\left[u_{j}^{2}u_{j^{\prime}}^{2}\right]$ \\
\midrule
$i=i^{\prime},j=j^{\prime}$ & $n^{2}$ & 3 & $\frac{k}{n}$ & $\frac{1}{2}$ & $\frac{3}{n(n+2)}$ \\
$i\neq i^{\prime},j=j^{\prime}$ & $n^{2}(n-1)$ & 1 & $\left(\frac{k}{n}\right)^{2}$ & $\frac{1}{2}$ & $\frac{3}{n(n+2)}$ \\
$i=i^{\prime},j\neq j^{\prime}$ & $n^{2}(n-1)$ & 3 & $\frac{k}{n}\cdot\frac{k-1}{n-1}$ & $\left(\frac{1}{2}\right)^{2}$ & $\frac{1}{n(n+2)}$ \\
$i\neq i^{\prime},j\neq j^{\prime}$ & $n^{2}(n-1)^{2}$ & 1 & $\left(\frac{k}{n}\right)^{2}$ & $\left(\frac{1}{2}\right)^{2}$ & $\frac{1}{n(n+2)}$ \\
\bottomrule
\end{tabular}
\caption{Overview of terms for Constant-fan-in type sparsity.}
\label{tab:tab3}
\end{table}

Which leads to the stated result.
\end{proof}
\begin{equation}
\end{equation}
\newpage
\section{Wide ResNet-22 trained on CIFAR-10}\label{sec:wide_res_net}

  \begin{figure}[tbp]
    \centering
    \includegraphics[width=0.46\linewidth]{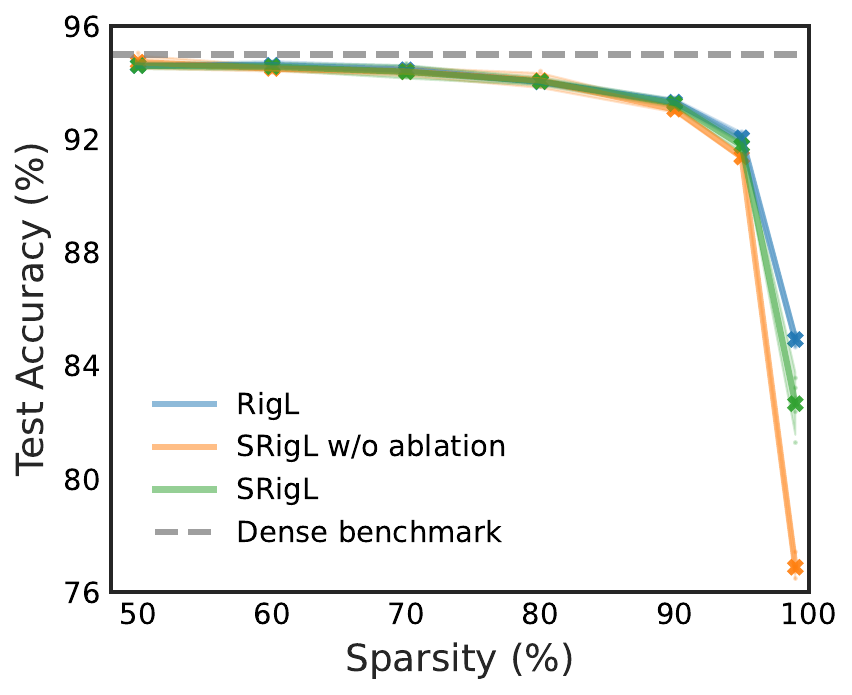}
     \qquad
    \renewcommand{\arraystretch}{0.5}
    \begin{tabular}[b]{@{}p{2em}cccc@{}}
    \toprule
    & \gls{rigl} & \multicolumn{2}{c}{\gls{srigl}} \\ 
    \cmidrule(lr){2-2 }\cmidrule(lr){3-4}
    sparsity (\%) &  & w/o & w/ ablation \\ 
    \midrule
    50 & $94.6\pm 0.1$ & $94.7\pm 0.1$ & $94.6\pm 0.1$ \\ \addlinespace
    60 & $94.6\pm 0.1$ & $94.5\pm 0.1$ & $94.6\pm 0.1$ \\ \addlinespace
    70 & $94.5\pm 0.1$ & $94.4\pm 0.1$ & $94.4\pm 0.1$ \\ \addlinespace
    80 & $94.0\pm 0.1$ & $94.1\pm 0.2$ & $94.0\pm 0.1$ \\ \addlinespace
    90 & $93.3\pm 0.1$ & $93.1\pm 0.1$ & $93.3\pm 0.1$ \\ \addlinespace
    95 & $92.1\pm 0.1$ & $91.4\pm 0.1$ & $91.8\pm 0.2$ \\ \addlinespace
    99 & $\mathbf{84.9\pm 0.2}$ & $76.9\pm 0.3$ & $82.7\pm 0.8$ \\ \addlinespace
    \midrule 
    0 & \multicolumn{3}{c}{\emph{dense Wide ResNet-22}:\quad $95.0$}\\ 
    \bottomrule 
    \end{tabular} 
    \captionlistentry[table]{A table beside a figure}
    \captionsetup{labelformat=andtable}
    \caption{Test accuracy of Wide ResNet-22 trained on CIFAR-10. Mean and 95\% confidence intervals are reported over five runs.}\label{table:wide_resnet22_table}
  \end{figure}

In \cref{table:wide_resnet22_table} we present results of training Wide ResNet-22~\citep{zagoruyko_wide_2017} with \gls{rigl} or \gls{srigl} on the CIFAR-10 dataset. The training details for this experiment are identical to those reported in \cref{sec:resnet18}. \gls{srigl} without ablation performs poorly at very high sparsities. With ablation, \gls{srigl} achieves generalization performance comparable to \gls{rigl}. 
\section{Hyperparameter and training details}\label{sec:training_details}
\subsection{ResNet-18 trained on CIFAR-10}\label{sec:cifar_training_details}
As per \citet{kuangliu_pytorch_cifar}, we modify the original ResNet-18 network by changing the kernel dimensions of the first convolutional layer to 3$\times$3 instead of 7$\times$7. Further, we reduce the stride in the first two convolutional layers to one to avoid excessive reduction of the feature map's spatial dimensions. 

We train each network for 250 epochs (97,656 steps) using a batch size of 128. An initial learning rate of 0.1 is reduced by a factor of 5 every 77 epochs (about 30,000 steps). We use stochastic gradient descent (SGD) with momentum,  with an L2 weight decay coefficient of 5e-4 and momentum coefficient of 0.9. We train each model using a single Nvidia V100 \gls{gpu}.

We achieve the desired overall sparsity by distributing the per-layer sparsity according to the \gls{erk}~\citep{evci_rigging_2021, mocanu_scalable_2018} distribution, 
which scales the per-layer sparsity based on the number of neurons and the dimensions of the convolutional kernel, if present. 
We set the number of mini-batch steps between connectivity updates, $\Delta T$, to 100. $\gamma_{sal}$ is set at 30\% based on the results of a small grid search performed on CIFAR-10 with ResNet-18 and Wide ResNet-22. See \cref{fig:min-sal-sweep} for details.

For each trial, we select a desired sparsity in the range from 0.5 to 0.99. At each connectivity update, the portion of weights to be pruned or regrown is based on a cosine annealing schedule~\citep{dettmers_sparse_2019} with an initial value $\alpha=0.3$. The portion of weights to be updated decays from the initial value to zero once 75\% of the total training steps have been completed, after which the weight mask remains constant.

\subsection{ResNet-50 trained on Imagenet}\label{sec:resnet50_training_details}
We use a mini-batch size of 512 instead of 4096,  We linearly scale the learning rate and $\Delta T$ to account for our smaller batch size. Linearly scaling the learning rate in this manner was included in the original \gls{rigl} source code and is further motivated by \citet{goyal2018}. We increase $\Delta T$ to 800 and average the dense gradients over eight mini-batch steps to ensure that \gls{srigl} has the same quality of parameter saliency information available as \gls{rigl} at each network connectivity update. We set $\gamma_{sal}$ to 30\% based on our grid search presented in \cref{fig:min-sal-sweep}.

Our learning rate uses a linear warm-up to reach a maximum value of 0.2 at epoch five and is reduced by a factor of 10 at epochs 30, 70, and 90. Using a mini-batch of 512, we train the networks for 256,000 steps to match \gls{rigl}'s training duration.  We use a cosine connectivity update schedule with $\alpha=0.3$. We initialize the sparse model weights per \citet{yani2022}. We train the networks using SGD with momentum, L2 weight decay, and label smoothing \citep{szegedy_rethinking_2016} coefficients of 0.9, 1e-4 and 0.1, respectively.

We use the same standard data augmentation in our data preprocessing as \gls{rigl}, including randomly resizing to 256$\times$256 or 480$\times$480 pixels, random crops to 224$\times$224 pixels, random horizontal flips, and per-image normalization to zero mean and unit variance using identical per RGB channel mean and standard deviation values as \gls{rigl}.  We train each model using either four Nvidia V100 or A100 \gls{gpu}s.

\subsection{Vision Transformer trained on ImageNet}\label{sec:vit_training_details}
For our \gls{vit} experiments, we used sparsity on the convolutional projection (input projection to patches), the fully connected layers in the feed forward (MLP) blocks and the output projections of the multi-headed attention (MHA) modules. We performed a lightweight ablation study on four \gls{vit} networks trained on ImageNet to determine the affect of sparsifying the first convolutional projection layer as well as the input projection layers in the multi-headed attention modules. Based on the results of our ablation study, \emph{we did not use sparsity on the MHA input projection layers or the scaled-dot products}. See \cref{fig:vit-mha-ablation} for more details.
This setup is similar to the "Sparse FF" models investigated by \citet{jaszczur_sparse_2021}. The global model sparsity level reported in \cref{table:vit_imagenet_table} is calculated based on the sparse modules only. If we also consider the parameters in the MHA input projections as part of our parameter budget, the global model sparsities tabulated in \cref{table:vit_imagenet_table} correspond to 60.35\% and 67.90\% for the rows labelled 80\% and 90\% sparsity, respectively. 

We add additional data augmentations following the standard TorchVision \citep{torchvision2016} \gls{vit} training procedure for ImageNet. These data augmentations applied include: random cropping, resizing the cropped image to 224 by 224 pixels, randomly horizontal flips, randomly augmenting with RandAugment algorithm \citep{cubuk_randaugment_2020}, and normalizing with the typical RGB channel mean and standard deviation values. We also randomly choose one of random mixup \citep{zhang_mixup_2023} or random cutmix \citep{yun_cutmix_2019} and add it to the above-noted augmentations. We use 0.2 and 1.0 for the alpha parameter values for mixup and cutmix, respectively. 
We omit Dropout \citep{srivastava_dropout_2014} from the model entirely to avoid potential layer collapse in the case where all non-zero weights are dropped from a layer and to avoid any other unintended interference with \gls{srigl}'s sparse training procedure. 

We sample eight mini-batch steps with 512 samples per mini-batch and accumulate gradients before applying the optimizer, resulting in an effective mini-batch size of 4096. We train the model for 150 epochs using an AdamW \citep{loshchilov_decoupled_2018} optimizer with weight decay, label smoothing, $\beta_1$, and $\beta_2$ coefficients of 0.3, 0.11, 0.9, and 0.999, respectively. We use cosine annealing with linear warm-up for our learning rate scheduler with an initial learning rate of $9.9\text{e-}5$ that warms-up to a maximum value of 0.003 at epoch 16. We clip all parameter gradients to a max L2 norm of 1.0. We apply uniformly distributed sparsity across all layers in the model. $\Delta T$ is set to 100 to update network connectivity every 100 mini-batch steps. We train each model using either four Nvidia V100 or A100 \glspl{gpu}.

\subsection{MobileNet-V3 trained on ImageNet}\label{sec:mobilenet}
We follow the TorchVision~\citep{torchvision2016} training recipe for MobileNet-V3 Large and Small for ImageNet. We set $\Delta T$ to 100 and $\gamma_{sal}$ to 30\% similar to our other \gls{cnn} experiments. We train the models from scratch for 600 epochs using an RMSProp~\citep{tieleman2012lecture} optimizer with a momentum, L2 weight decay, and smoothing constant coefficients of 0.9, 1e-5, and 0.9, respectively. The networks are trained with a step learning rate decay schedule with initial learning rate of 0.064, multiplicative factor of 0.973, and we decay the learning rate every two epochs. 

The input data is augmented with random cropping to 224 by 224 pixels, random horizontal flips, AutoAugmentation using the ImageNet policy~\citep{cubuk_autoaugment_2019}, normalizing to standard RGB mean and standard deviation values, and random erasing with a probability of 0.2 \citep{zhong_random_2017}. Similar to the above, we omit Dropout~\citep{srivastava_dropout_2014} to avoid potential layer collapse. Unlike the TorchVision recipe, we \emph{do not} average the trained parameters across the last three checkpoints that improved the top-1 accuracy. 
We train with a batch size of 512 and accumulate gradients across two mini-batches, resulting in an effective mini-batch size of 1024. We train each model using four Nvidia A100 \glspl{gpu}. 

\begin{center}
\begin{figure}[tbph]
\begin{minipage}{0.47\linewidth}
        \centering
        \includegraphics[width=\linewidth]{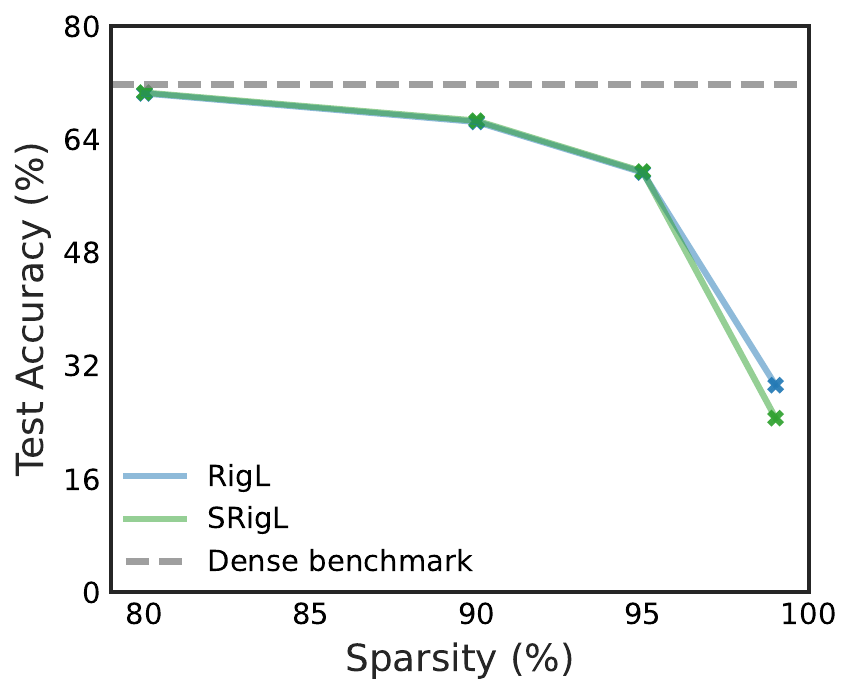}
        \caption{\textbf{MobileNet-V3 Large / ImageNet} top-1 accuracy. \gls{srigl} compares well against \gls{rigl} both both models perform poorly compared to the denes baseline at 99\% sparsity.}
        \label{fig:mobilenet-large}
\end{minipage}
\hfill
\begin{minipage}{0.47\linewidth}
        \includegraphics[width=\linewidth]{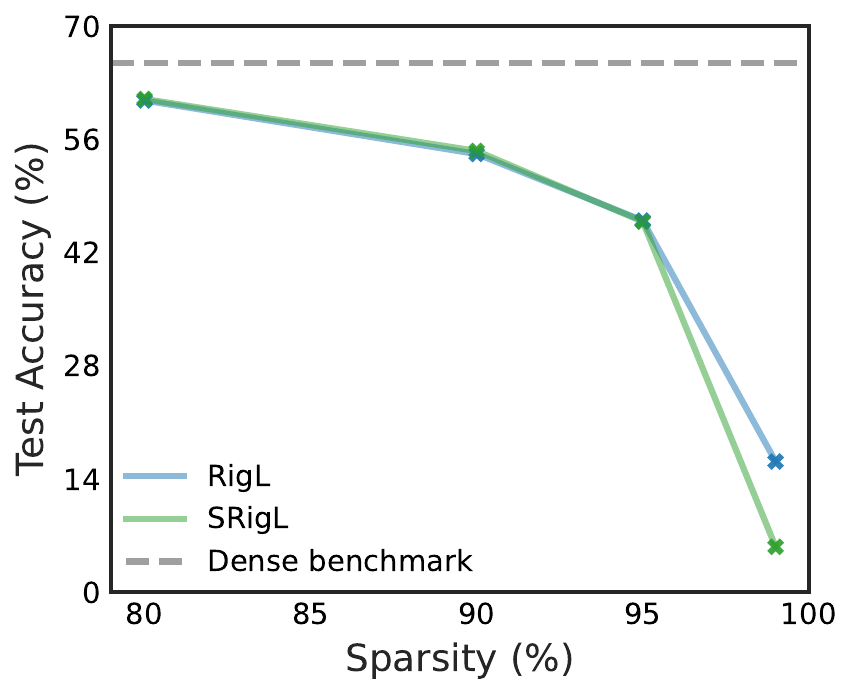}
        \caption{\textbf{MobileNet-V3 Small / ImageNet} top-1 accuracy. \gls{srigl} compares well against \gls{rigl} both both models perform poorly compared to the denes baseline at 99\% sparsity.}
        \label{fig:mobilenet-small}
\end{minipage}
\end{figure}
\end{center}
\section{Tuning \texorpdfstring{$\gamma_{sal}$}{ablation saliency hyperparameter}, minimum percentage salient weights per neuron}\label{sec:min_sal_weights_sweep}
\begin{figure*}[ht]
    \centering
    \begin{subfigure}{0.49\linewidth}
    \includegraphics[width=\linewidth]{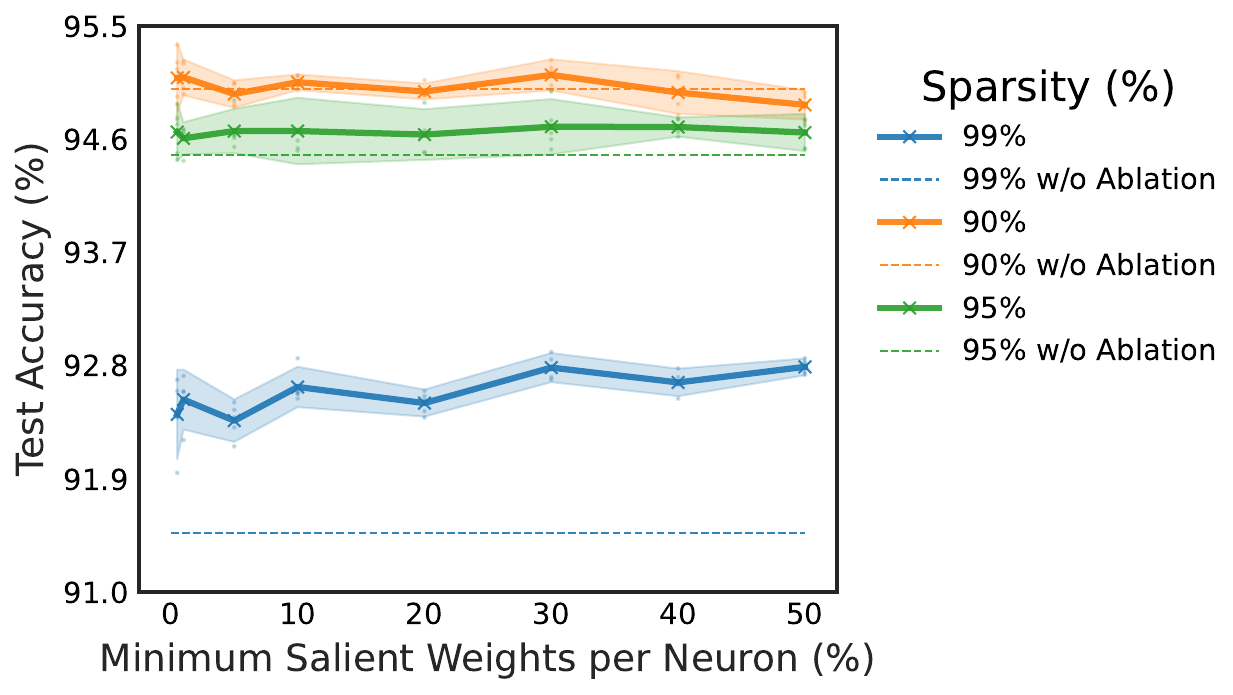}
    \caption{ResNet-18/CIFAR-10}\label{fig:resnet18_salient_sweep}
    \end{subfigure}
    \hfill
    \begin{subfigure}{0.49\linewidth}
    \includegraphics[width=\linewidth]{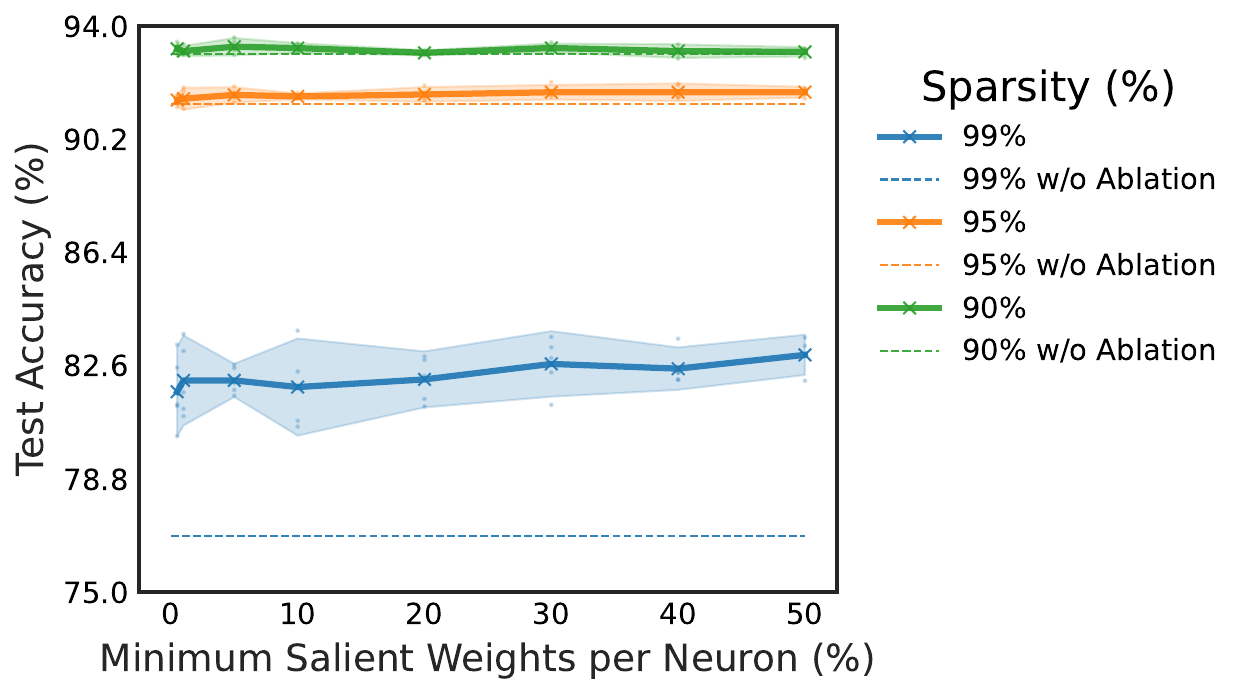}
    \caption{Wide ResNet-22/CIFAR-10}\label{fig:wide_resnet22_salient_sweep}
    \end{subfigure}
    \hfill
    \caption{\textbf{(\subref{fig:resnet18_salient_sweep}) \textbf{ResNet18/CIFAR-10} Test Accuracy vs. $\gamma_{sal}$} when trained with \Gls{srigl} with and without ablation for a range of sparsities. The mean and 95\% confidence intervals are shown for five different random seeds for the runs with ablation. For the runs without ablation, we report the mean of five different random seeds. \textbf{(\subref{fig:wide_resnet22_salient_sweep}) Wide ResNet-22 Test Accuracy vs. $\gamma_{sal}$}. The mean and 95\% confidence intervals are shown for five different random seeds.
    }\label{fig:min-sal-sweep}
\end{figure*}

\begin{figure}[tp]
\begin{minipage}{0.47\linewidth}
    \includegraphics[width=\linewidth]{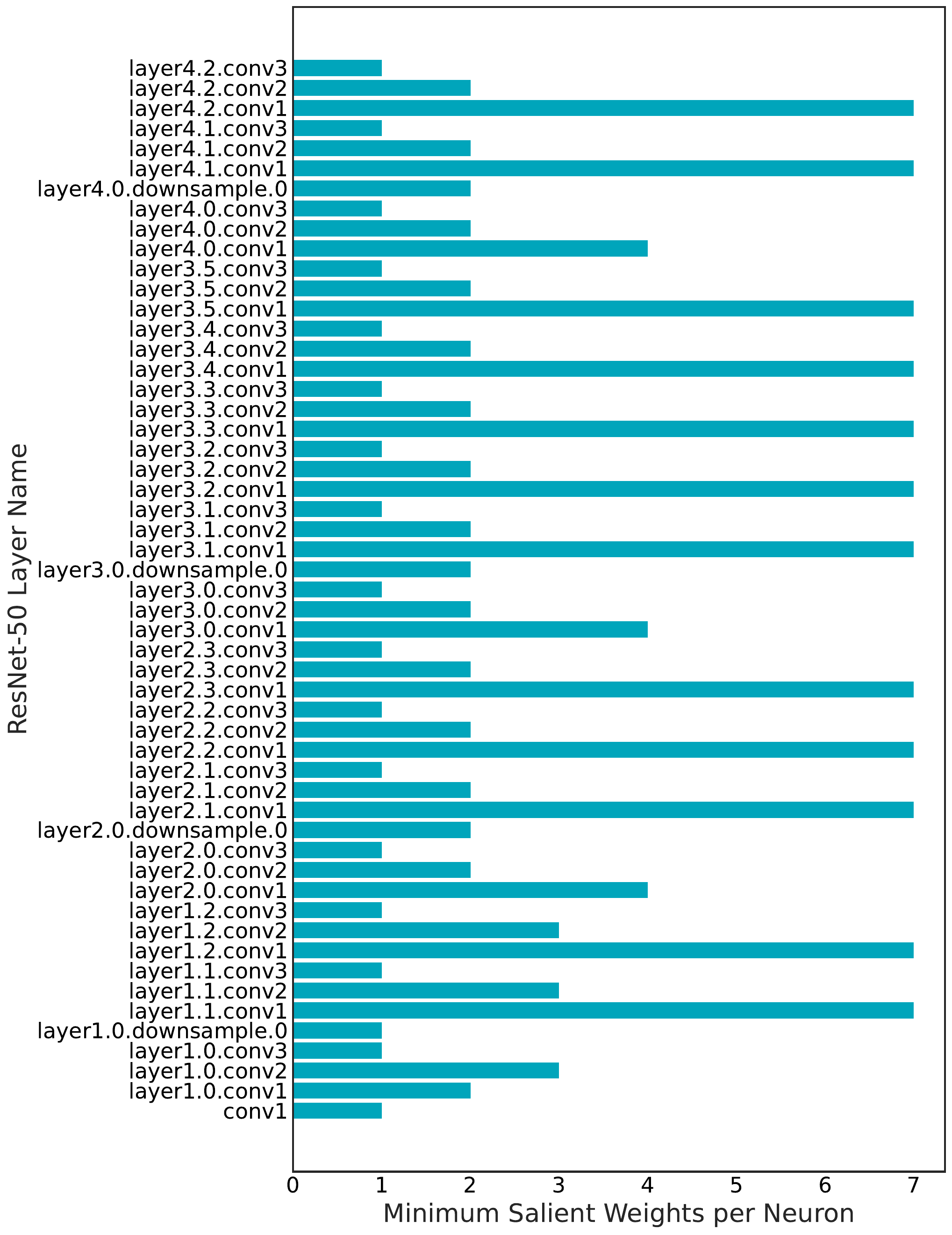}
    \caption{\textbf{ResNet-50 Layer vs. Minimum salient weights per neuron}. \gls{srigl} sets the minimum salient weight per neuron to 1 if the product between $\gamma_{sal}$ and the sparse fan-in per neuron is less than 1. Therefore, even in a relatively large network such as ResNet50 many of the layers only require that a single weight be active to keep the neuron active. We believe this is why \gls{srigl}'s performance is relatively invariant to various ablation thresholds when applied to \glspl{cnn}}\label{fig:layer-vs-min-sal}
    \end{minipage}
    \hfill
    \begin{minipage}{0.47\linewidth}
    \includegraphics[width=\linewidth]{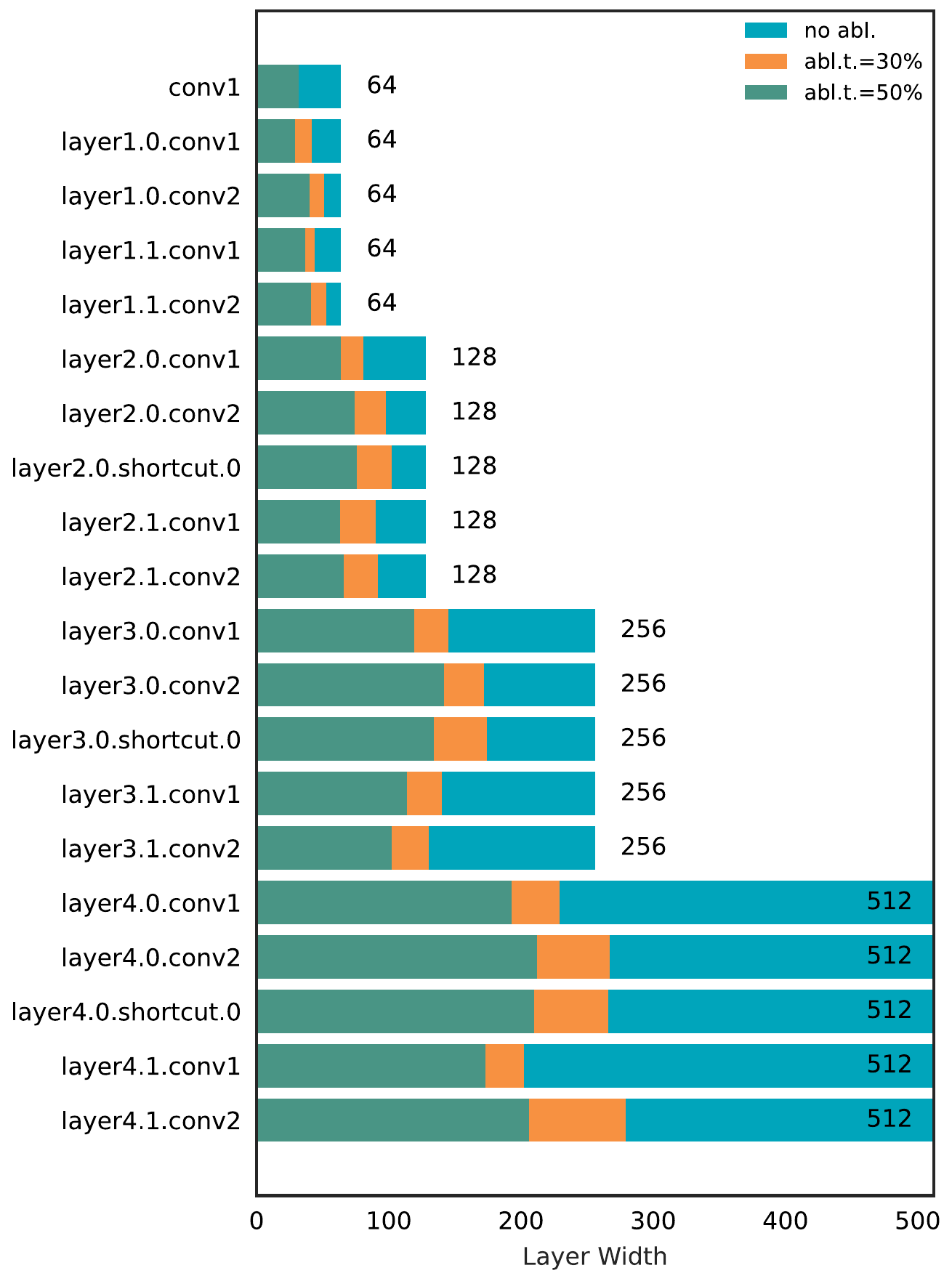}
    \caption{\textbf{ResNet-18/CIFAR-10 layer widths at the end of training at 99\% sparsity}. Without ablation, constant fan-in constraint enforces that sparse layers retain their original width. When ablation is enabled, the $\gamma_{sal}$ threshold (minimum percentage salient weights per neuron) is used to control the amount of ablation. 
    }\label{fig:resnet18_width}
\end{minipage}
\end{figure}

\cref{fig:min-sal-sweep} depicts the generalization performance of highly sparse ResNet-18 and Wide ResNet-22 models trained on the CIFAR-10 dataset. \gls{srigl}'s generalization performance at high sparsities is improved with neuron ablation; however, the specific value selected for $\gamma_{sal}$ does not have a significant effect on performance. Our experiments demonstrate that \gls{srigl} performs well with a variety of $\gamma_{sal}$ values. In \cref{sec:results} we report the results of \gls{srigl} models trained with $\gamma_{sal}$ set to 30\%. With dynamic ablation enabled, we set the minimum salient weights per neuron to one if the user-defined threshold results in a value less than one. In \cref{fig:layer-vs-min-sal}, many layers in ResNet-50 are set to the minimum threshold of one when we apply a $\gamma_{sal}$ of 30\% for all model types other than \gls{vit}. This minimum threshold explains the invariance of the model's performance when comparing against multiple values for $\gamma_{sal}$. 

\begin{figure*}[ht]
    \centering
    \begin{subfigure}[t]{0.49\linewidth}
    \includegraphics[width=\linewidth]{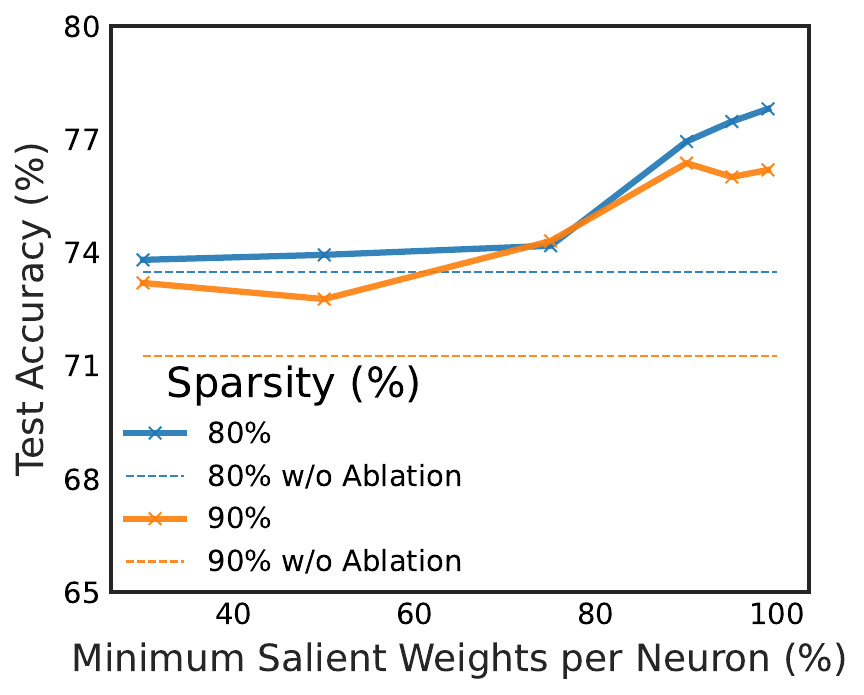}
    \caption{\textbf{\gls{vit}/ImageNet Test Accuracy vs. $\gamma_{sal}$} when trained with \gls{srigl} with and without ablation enabled for 80\% and 90\% sparsity. \gls{vit}'s performance is much more sensitive to $\gamma_{sal}$ and generally performs best with high ablation thresholds. Based on this data we set $\gamma_{sal}$ to 95\% for our results reported in \cref{sec:vit}.}\label{fig:vit_min_sal}
    \end{subfigure}
    \hfill
    \begin{subfigure}[t]{0.49\linewidth}
    \includegraphics[width=\linewidth]{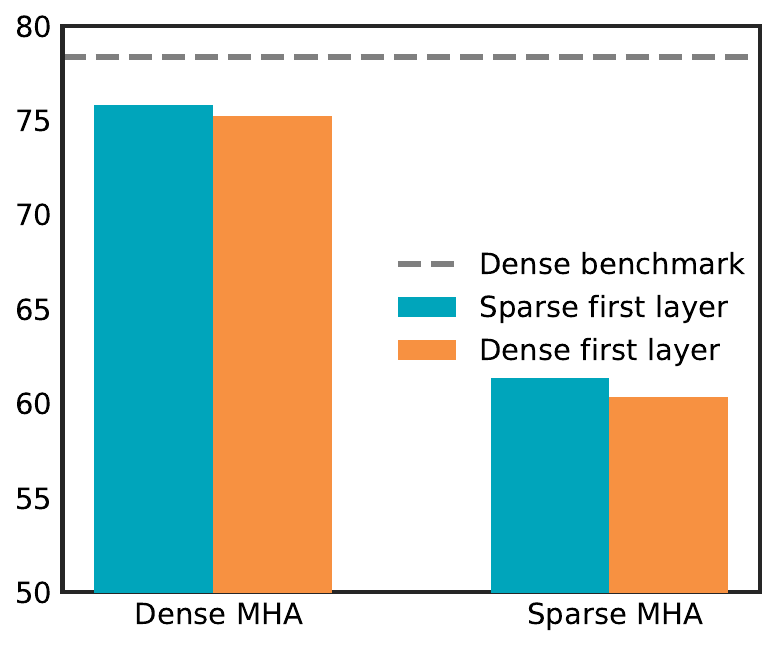}
    \caption{\textbf{\gls{vit} ablation study}. The best performing variant used a sparse first layer and dense input projections in the MHA modules.}
    \label{fig:vit-mha-ablation}
    \end{subfigure}
\end{figure*}

\cref{fig:vit_min_sal} demonstrates how \gls{vit}'s generalization performance is much more sensitive to $\gamma_{sal}$. We find that \gls{rigl} learns a sparse connectivity pattern with a large variance in sparse fan-in between neurons within a given layer, with some neurons having an order of magnitude more fan-in connection than the mean fan-in. 

\begin{figure}[bp]
\begin{minipage}{0.47\linewidth}
    \includegraphics[width=\linewidth]{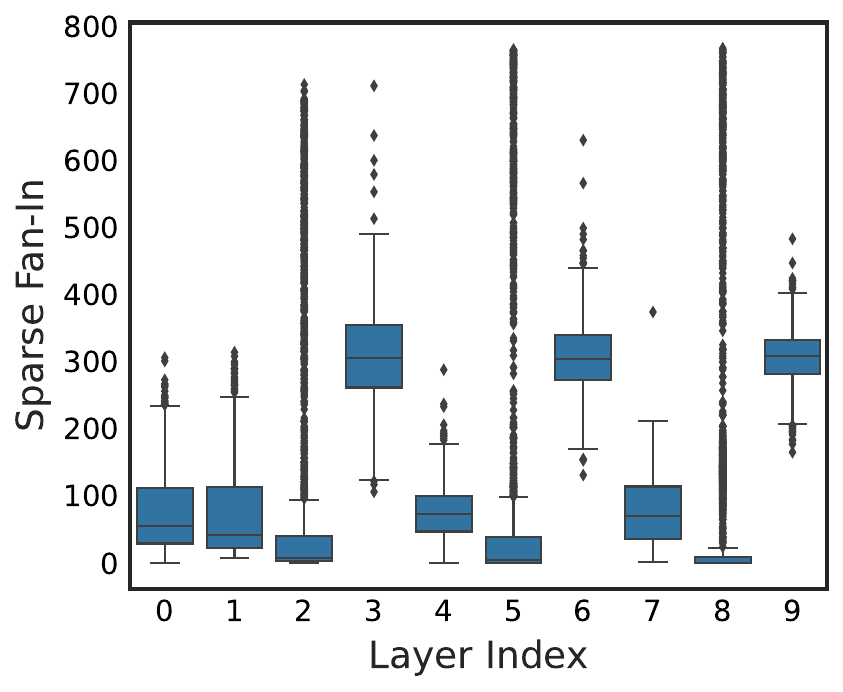}
    \caption{\textbf{Sparse Fan-In vs. \gls{vit} layer index at the end of training with \gls{rigl} at 90\% sparsity}. Only the first 10 layers are shown for clarity. We find that \gls{rigl} learns a sparse connectivity with large variance in fan-in between neurons within the same layer with some neurons receiving up to $\times 10$ the number of active connections than the mean for the same layer.}\label{fig:vit-rigl-fan-in}
    \end{minipage}
    \hfill
    \begin{minipage}{0.47\linewidth}
        \includegraphics[width=\linewidth]{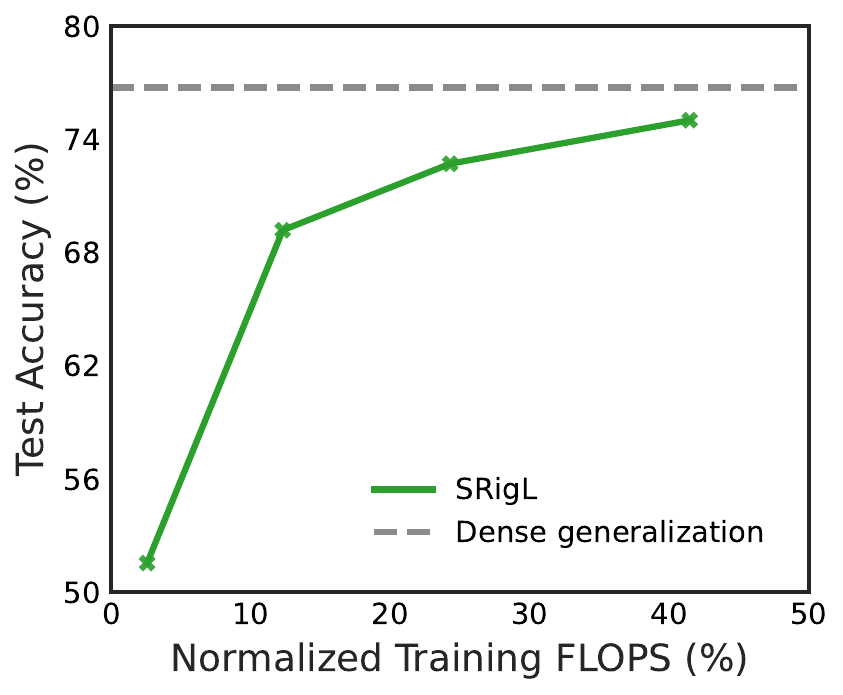}
        \caption{\textbf{Training \gls{flops}} for \gls{srigl} on ResNet-50/ImageNet at a variety of sparsities compared with dense generalization. \gls{flops} are normalized by dense training \gls{flops}.}\label{fig:combined-flops}
    \end{minipage}
\end{figure}

\section{Condensed matrix multiplication}\label{sec:condensedrepresentationmath}
Using a constant fan-in sparse representation presents an advantage compared to the general N:M sparse representation in that we can represent our weight matrices in a compact form, since every neuron/convolutional filter has the same number of non-zero weights. Here we demonstrate how this can be used to accelerate a fully-connected layer.

Consider the standard matrix-vector product:
\begin{align}\label{eq:Wv_general}\tiny
Wv = 
    \begin{pmatrix}
        W_{11}  & W_{12} & \hdots & W_{1d} \\
        W_{21}  & W_{22} & \hdots & W_{2d} \\
         \vdots & \vdots & \ddots & \vdots \\
        W_{n1}  & W_{n2} & \hdots & W_{nd}       
    \end{pmatrix}
    \begin{pmatrix}
        v_1 \\
        v_2 \\
        \vdots \\
        v_d       
    \end{pmatrix}
    =
    \begin{pmatrix}
        \sum\limits_{j=1}^d W_{1j} v_j \\
        \sum\limits_{j=1}^d W_{2j} v_j \\
        \vdots \\
        \sum\limits_{j=1}^d W_{nj} v_j \\  
    \end{pmatrix}
    = v^{\text{out}}
\end{align}

When $W\in\mathbb{R}^{n\times d}$ is sparse and has only $k$ non-zero elements per row, 
the sums representing each element of $v^\text{out}$ will be limited to $k$ terms, i.e.: 
\begin{equation}
    v^\text{out}_i = \sum\limits_{\alpha=1}^k  W_{i{j_\alpha}} v_{j_\alpha} \quad\text{ with }\,\, {j_\alpha\in\{1,\dots,d\}},\quad j_\alpha\neq j_{\alpha^\prime}
\end{equation}

Note that the expression on the right-hand side of Eq.~\eqref{eq:Wv_general} 
can be represented as an operation between 
a dense matrix $W^c \in \bR^{n\times k}$ (we call it ``condensed $W$'') 
and $k$ vectors $v^{\pi_1},\dots, v^{\pi_k}$, $v^{\pi_i}\in\bR^n$, 
whose elements are drawn from $v$ with replacement (we call them ``recombinations of $v$''). 
The operation is a sum over element-wise products between the $i$-th column of $W^c$ 
and the $i$-th column vector $v^{\pi_i}$:
\begin{equation}
    Wv = \sum\limits_{i=1}^{k} W^c_{:,i}\odot v^{\pi_i}
\end{equation}

Mathematically, these methods are equivalent for any matrices. 
Computationally, the condensed method can be more efficient, 
in particular for sparse matrices with constant small fan-in $k$. 
By construction, this method requires the sparse matrix $W$ to be stored in dense representation which involves two 2D arrays of shape $n\times k$: 
One holds the \textit{values} of the non-zero elements of $W$ and the other one their respective \textit{column indices}, which are used to generate input vector re-combinations. An efficient computational implementation of this method is subject of ongoing work on this project. Based on our results, the constant fan-in constraint does not appear to have a limiting effect on \glspl{snn}.

\section{\protect\glsentrytext{flops} analysis}\label{sec:flops}
In \cref{fig:combined-flops}, we present an analysis of the \gls{flops} required during training and inference for \gls{srigl} and compare with \gls{srste}. We calculate \gls{flops} using the same methodology as \citet{evci_rigging_2021} by considering only operations induced by convolutional and linear layers and their activations. \gls{flops} for add and pooling operations are ignored. For training \gls{flops}, we also disregard \gls{flops} required for mask updates, as this step is amortized over $\Delta T$ steps and is negligible compared to the \gls{flops} required otherwise for training. The open-source code for counting operations is from the NeurIPS 2019 MicroNet Challenge and is available on GitHub\footnote{\href{https://github.com/google-research/google-research/tree/master/micronet\_challenge}{MicroNet Challenge Github Repository}}.

Similar to other \gls{dst} methods, \gls{srigl} obtains generalization performance comparable to a dense network benchmark at a fraction of the \gls{flops} required for both training and inference. 
\section{In Time Overparameterization Rates}
\begin{figure}[h]
\begin{minipage}{0.47\linewidth}
    \includegraphics[width=\linewidth]{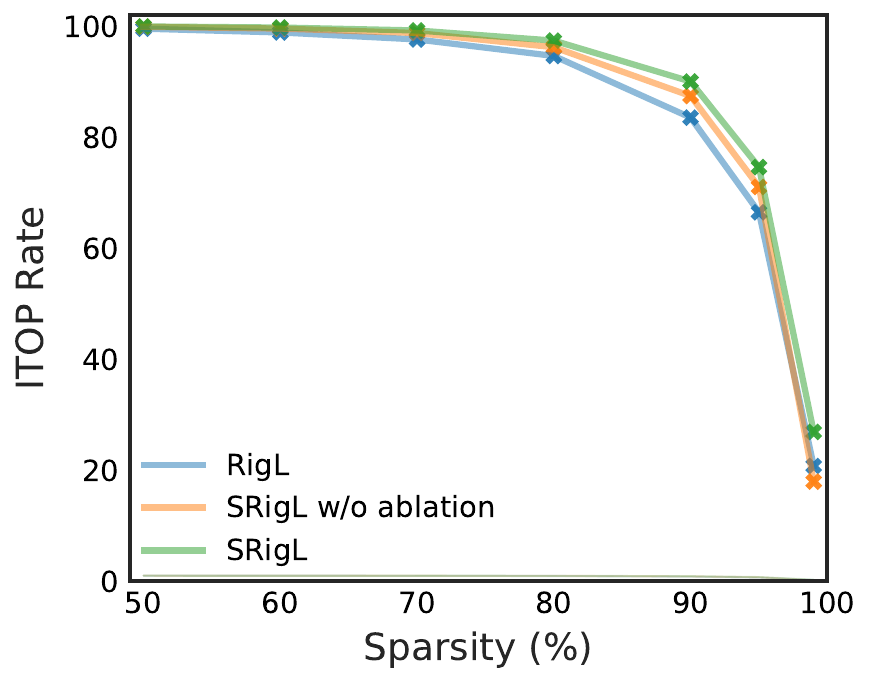}
    \caption{\textbf{ResNet-18/CIFAR-10 ITOP rate}}\label{fig:wideresnet-itop}
    \end{minipage}
    \hfill
    \begin{minipage}{0.47\linewidth}
    \includegraphics[width=\linewidth]{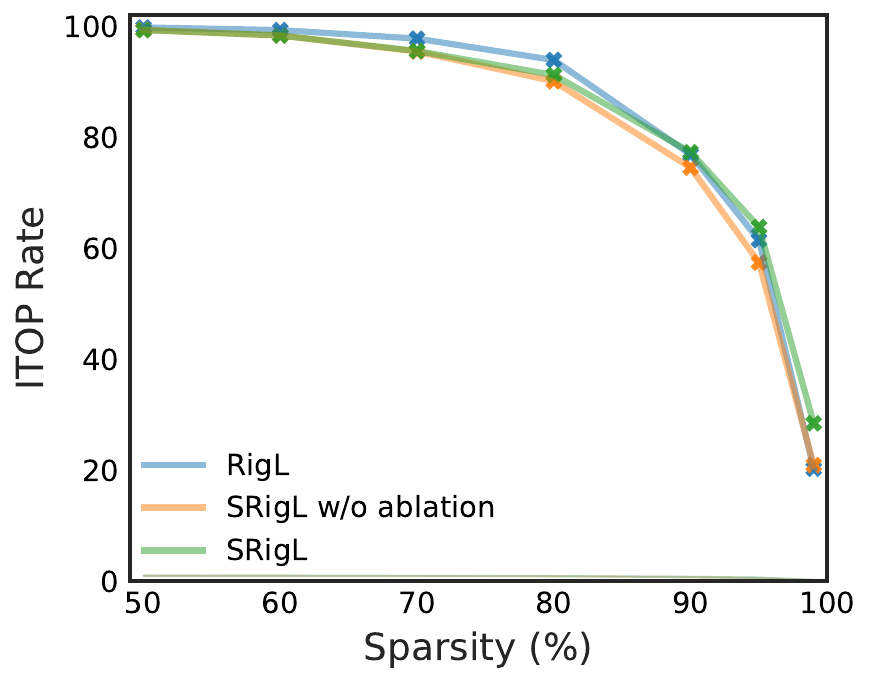}
    \caption{\textbf{ResNet-18/CIFAR-10 ITOP rate}}\label{fig:resnet18-itop}
\end{minipage}
\end{figure}

\begin{figure}[h]
\begin{minipage}{0.47\linewidth}
    \includegraphics[width=\linewidth]{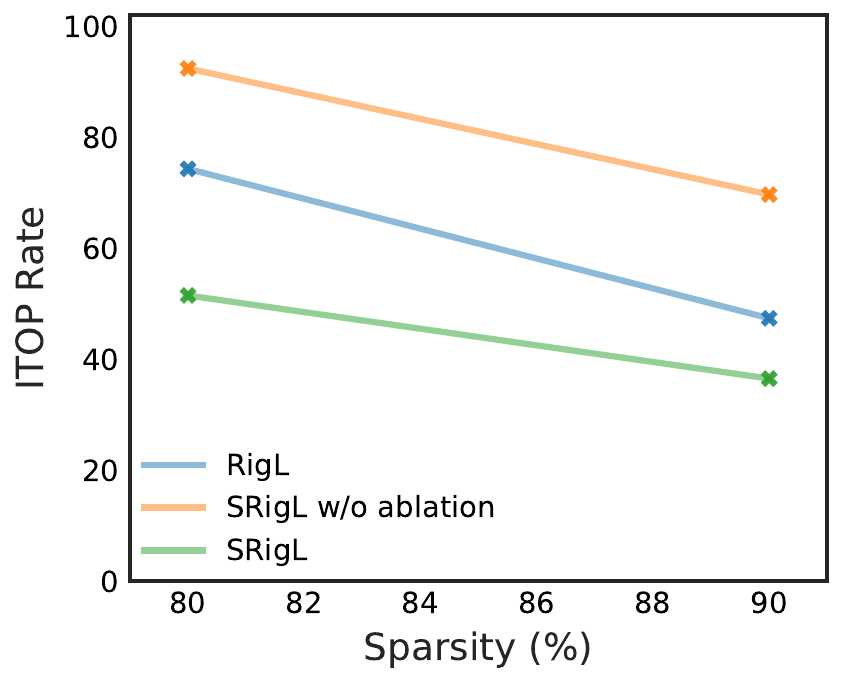}
    \caption{\textbf{\gls{vit}/ImageNet ITOP Rate}}\label{fig:vit-itop}
    \end{minipage}
    \hfill
    \begin{minipage}{0.47\linewidth}
        \includegraphics[width=\linewidth]{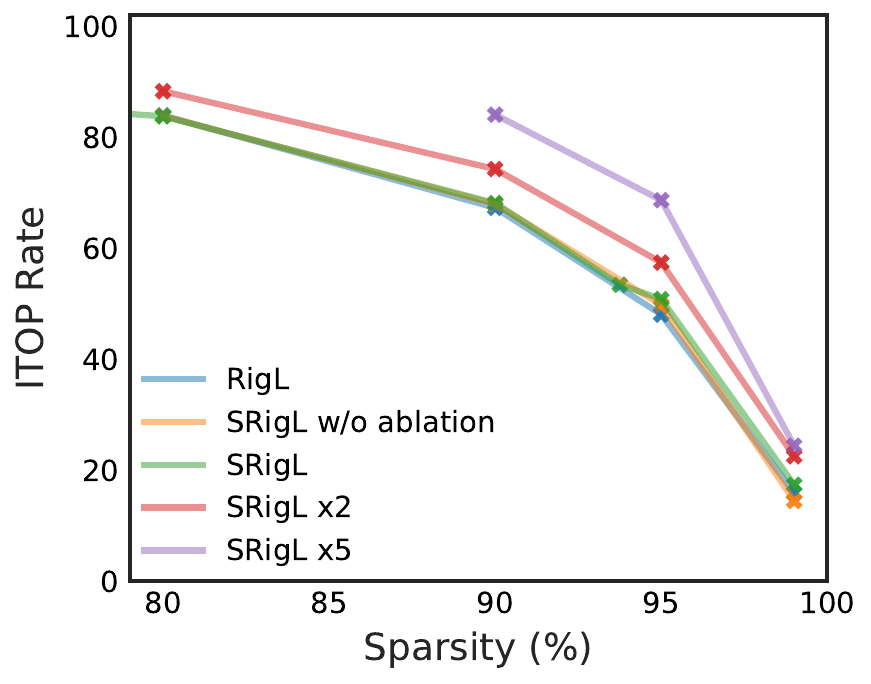}
        \caption{\textbf{ResNet-50/ImageNet ITOP Rate}}\label{fig:imagenet-itop}
    \end{minipage}
\end{figure}

In \cref{fig:vit-itop,fig:imagenet-itop,fig:resnet18-itop,fig:wideresnet-itop} we present the \gls{itop} \citep{liu_we_2021} for various models and datasets. In this same work, \citet{liu_we_2021} proposed modified hyperparameters for \gls{rigl} that may yield higher generalization performance; however, a detailed investigation of these hyperparameters for \gls{srigl} is left to future work. 

\newpage
\section{Condensed linear \glsentryshort{cpu} benchmark details} \label{sec:timingsdetails}
\begin{figure}[b!]
        \centering
        \includegraphics[width=\linewidth]{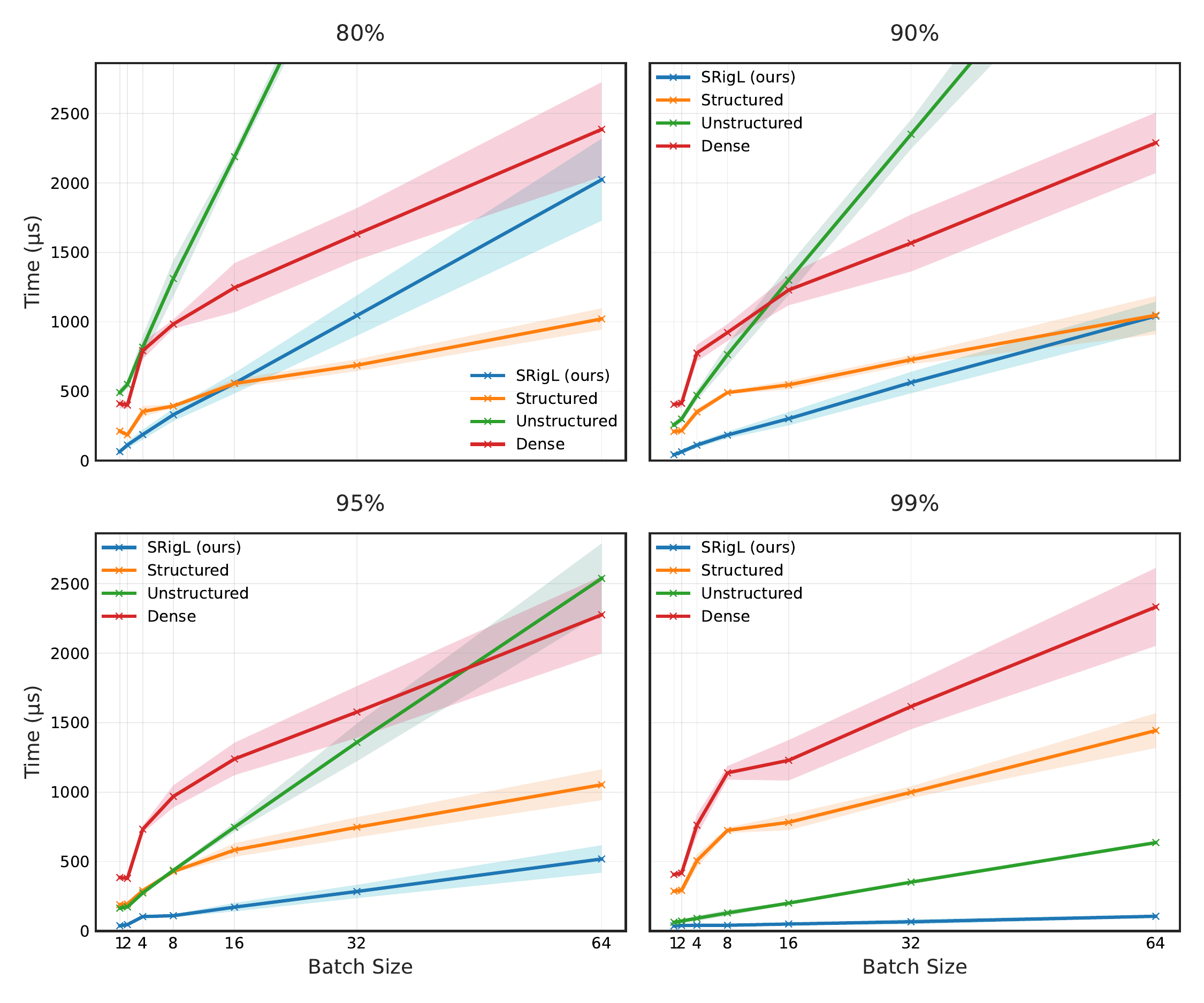}
        \caption{\gls{cpu} benchmarks with 1 thread up to batch size 64}
        \label{fig:thread-1-bs-64}
\end{figure}

\begin{figure}[tbp]
        \vspace{-1.5cm}
        \centering
        \includegraphics[width=\linewidth]{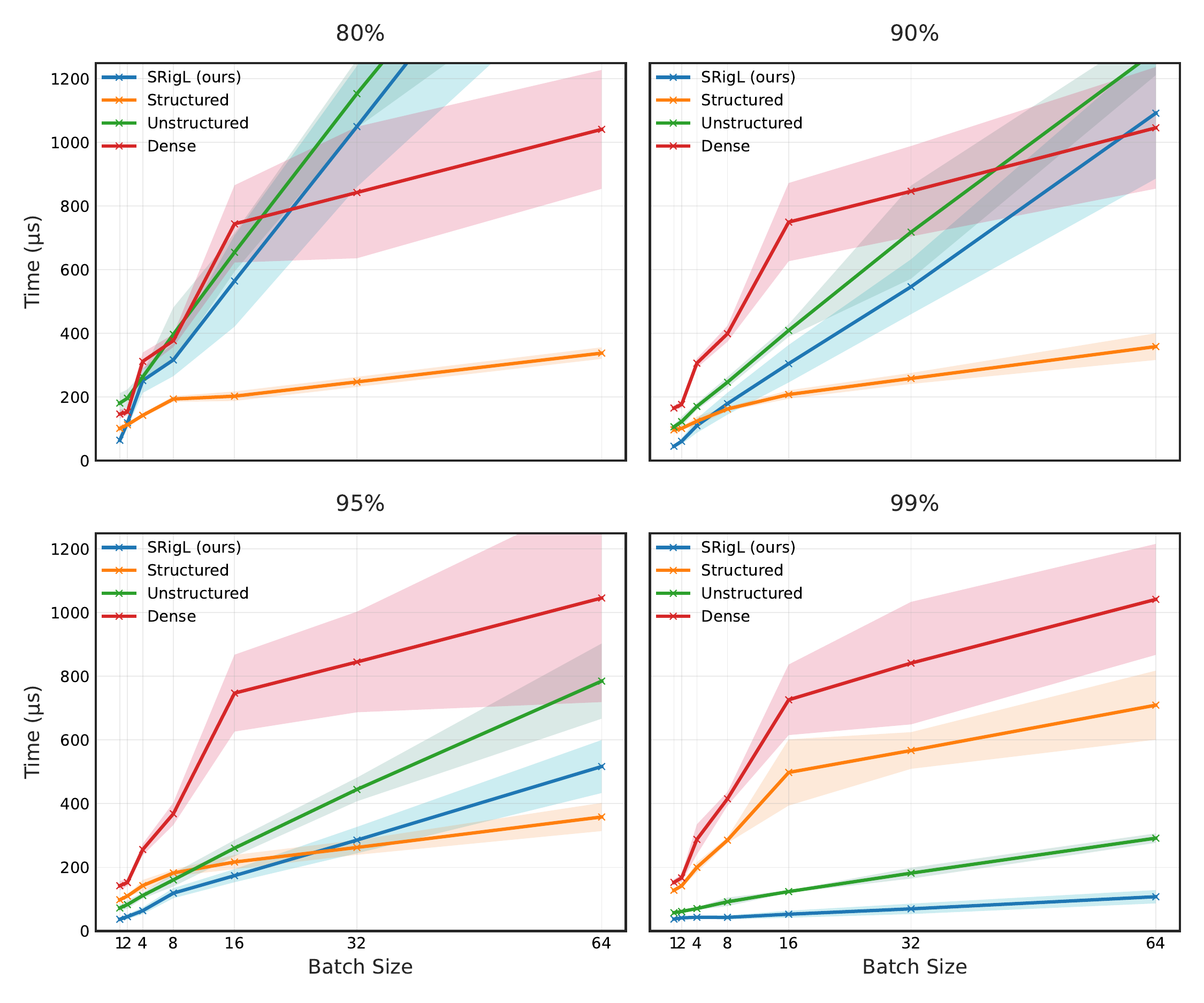}
        \caption{\gls{cpu} benchmarks with 4 threads up to batch size 64}
        \label{fig:thread-4-bs-64}
        \includegraphics[width=\linewidth]{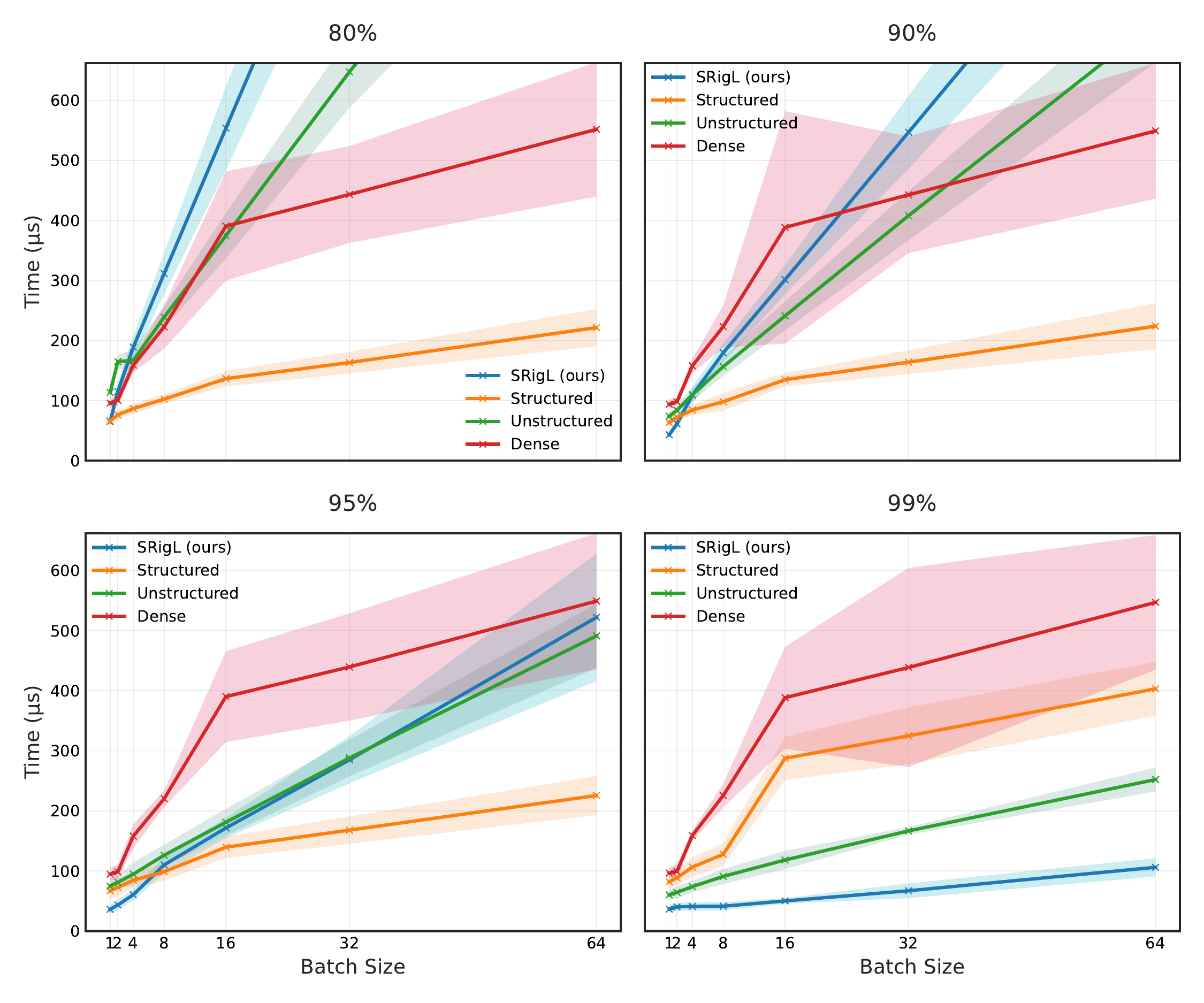}
        \caption{\gls{cpu} benchmarks with 8 threads up to batch size 64}
        \label{fig:thread-8-bs-64}
\end{figure}

For each sparsity level, we used the trained weights from the last linear layer in the final multi-layer perception block from the \gls{vit} transformer encoder. This layer has a width of 768 neurons and an input of 3072 features. The input and layer parameters are all set to a 32 bit floating point type. 
Across all sparsities, batch sizes, and number of threads investigated, our condensed representation utilizing both structured and fine-grained sparsity yields the fastest online inference speed. However, at higher batch sizes and modest sparsities, structured sparsity is often faster than our condensed representation. See \cref{fig:thread-1-bs-64,fig:thread-4-bs-64,fig:thread-8-bs-64} for benchmark results from 1-8 threads and batch sizes 1-64. We note that \gls{srigl} with either a condensed or a structured sparse representation yields the fastest benchmark times. 

We used \texttt{torch.compile} with the inductor backend. For compiler options, we used the max-autotune mode and full graph output. However, full graph output is not compatible with \gls{csr} formats so we omit this parameter for the unstructured benchmarks. The benchmark script was run with a \texttt{niceness} value of $-15$ to ensure as accurate results as possible. The apparent slow down in 99\% structured sparse benchmarks compared to other sparsities is due to the fact that \gls{srigl} ablates fewer neurons at 99\% sparsity. At extreme sparsities, each neuron has very few active weights resulting in more neurons being considered as \emph{salient} by \gls{srigl}.

\section{GPU Benchmarks}\label{sec:gpu-benchmarks}
\begin{figure*}[tbph]
        \centering
    \begin{subfigure}{0.95\linewidth}
        \includegraphics[width=\linewidth]{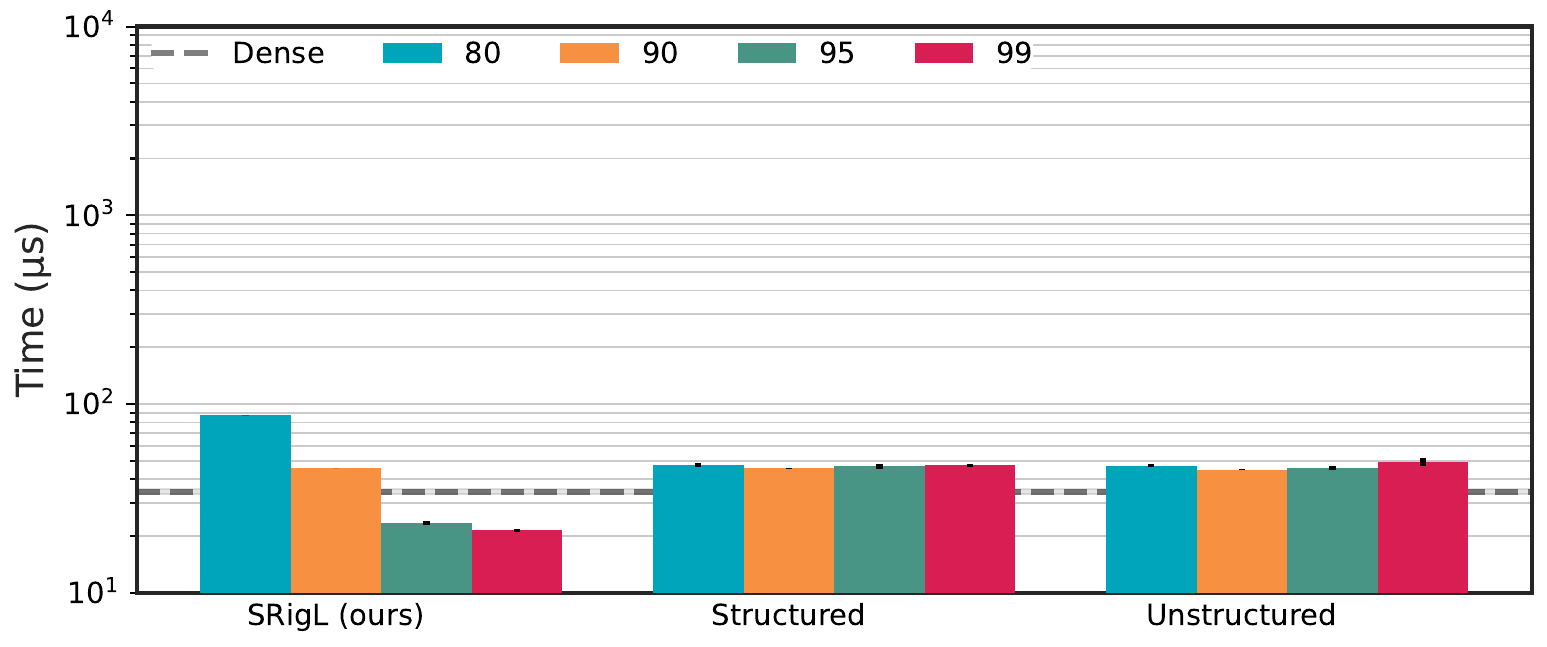}
        \caption{\textbf{\gls{gpu} online inference (batch size of 1)}}\label{fig:gpu_accel_bs_1}
    \end{subfigure}

    \begin{subfigure}{0.95\linewidth}
        \centering
        \includegraphics[width=\linewidth]{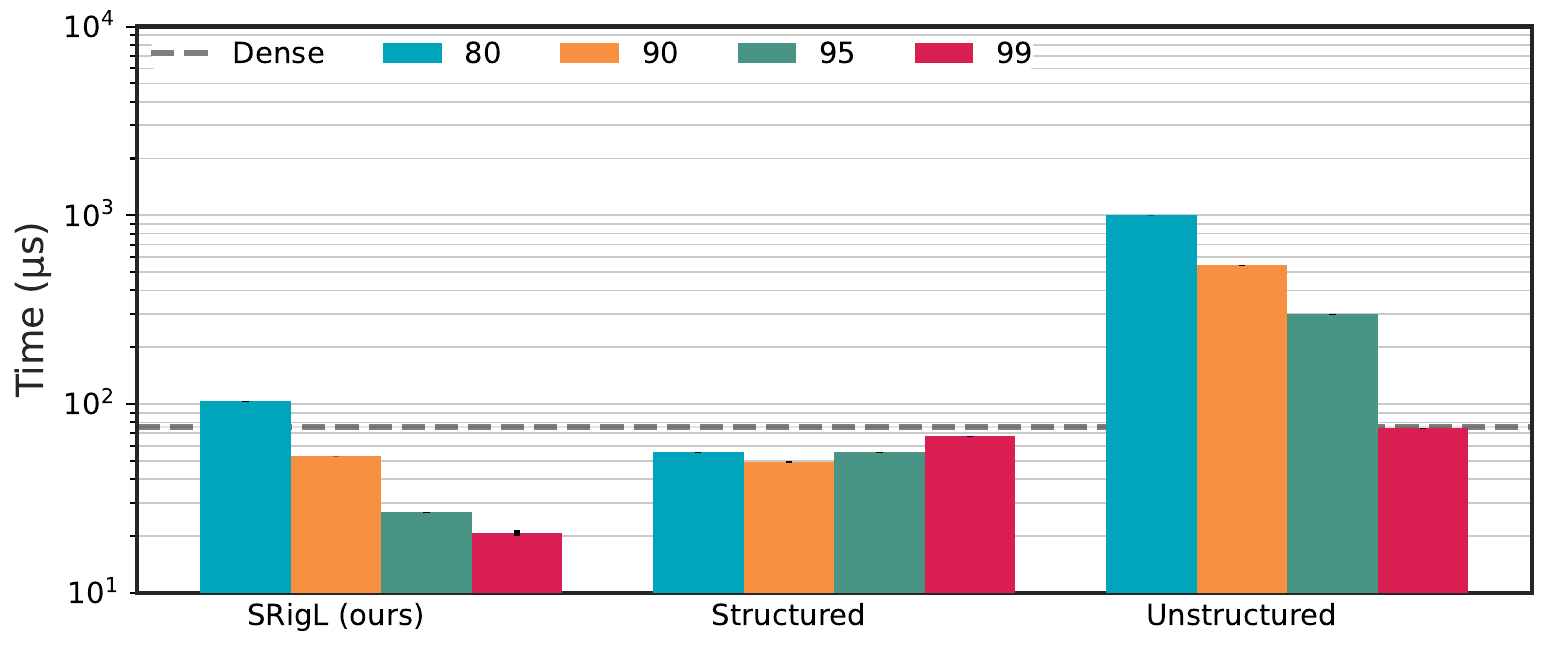}
        \caption{\textbf{\gls{gpu} batched inference with batch size of 128}}\label{fig:gpu-accel-bs-128}
    \end{subfigure}
    
    \begin{subfigure}{{0.95\linewidth}}
        \centering
        \includegraphics[width=\linewidth]{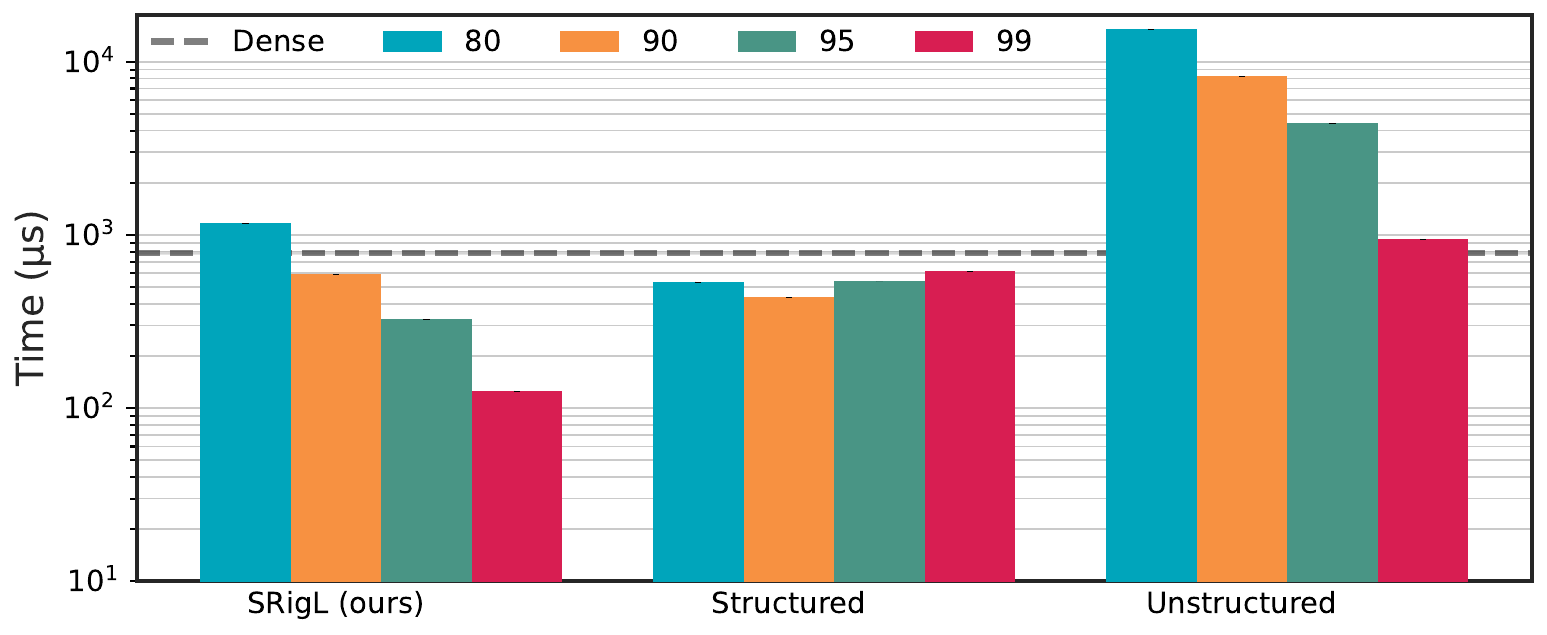}
        \caption{\textbf{\gls{gpu} batched inference with batch size of 2048}}\label{fig:gpu-accel-bs-2048}
    \end{subfigure}
\caption{\textbf{Real-world \gls{gpu} wall-clock timings for inference} on an NVIDIA Titan V. We compare timings for a fully-connected layer extracted from the \gls{vit} model trained with \gls{srigl} when compressed using the condensed representation learned by \gls{srigl}, structured (i.e.\ \gls{srigl} with only neuron ablation) and unstructured (i.e.\ \gls{csr}) representations. Batch sizes are 1, 256, and 2048 for sub-figures \ref{fig:gpu_accel_bs_1}, \ref{fig:gpu-accel-bs-128}, \ref{fig:gpu-accel-bs-2048}, respectively. The median over a minimum of 5 runs is shown, while the error bars show the std.\ dev. Note: y-axis scale is logarithmic}\label{fig:gpu-timings}
\end{figure*}

Using \gls{gpu} CUDA kernels developed by \citep{schultheis_towards_2023}, we accelerate our sparse networks and demonstrate a significant acceleration for batched inference and a modest acceleration for online inference at high sparsities (>90\%), see \cref{fig:gpu-timings}. All runs conducted on an NVIDIA Titan V. Note y-axis scale is logarithmic.

\FloatBarrier
\begin{figure}[th!]
        \centering
        \includegraphics[width=\linewidth]{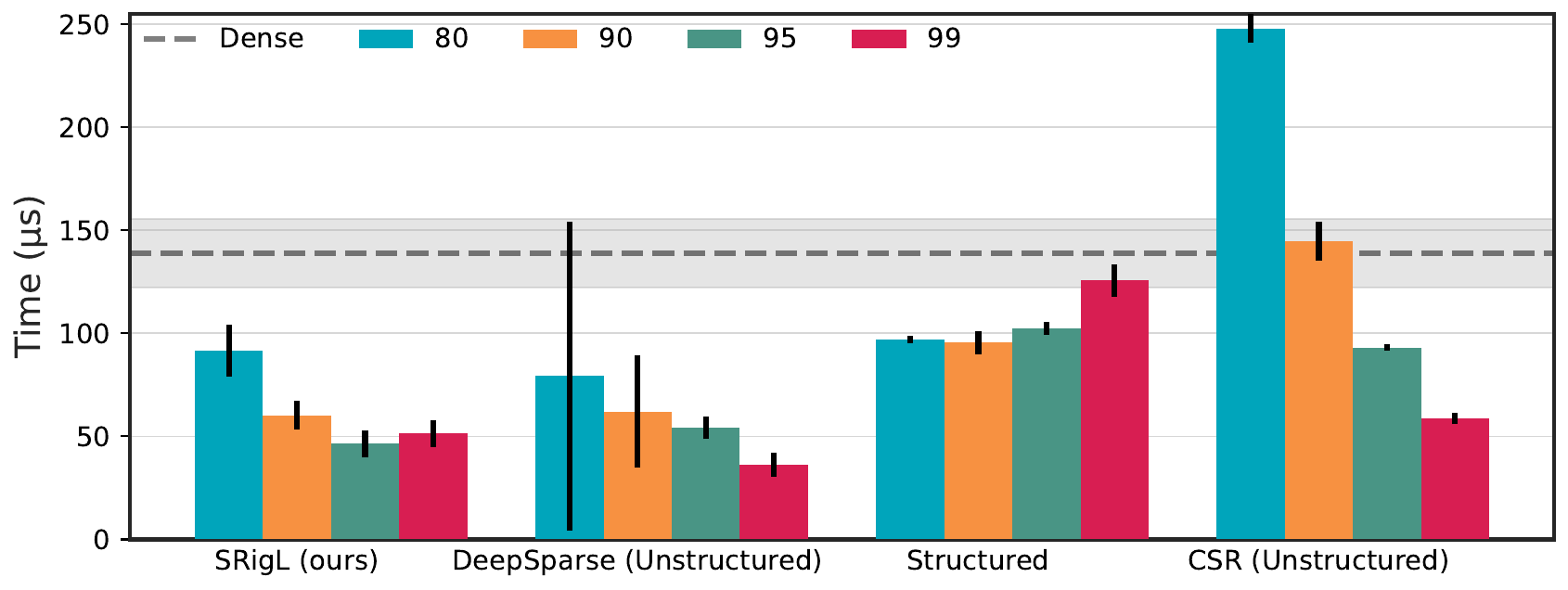}
        \caption{\textbf{Online inference with DeepSparse compared to \gls{srigl}} on an Intel Xeon W-2145 with 4 threads. The median over a minimum of 5 runs is shown, while the error bars show the std.\ dev.}
        \label{fig:deep-sparse-benchmark}
\end{figure}
\section{DeepSparse CPU Benchmarks}\label{sec:deepsparse}
Here we present online inference benchmarks for CPU using the DeepSparse Engine library \citep{DBLP:journals/corr/abs-2111-13445}. DeepSparse library includes several engineering innovations to accelerate unstructured sparsity on \gls{cpu}. For instance, a \emph{depth-wise asynchronous} execution algorithm is used that takes advantage of the relatively large cache size for \glspl{cpu} compared to hardware accelerators such as \glspl{gpu}. Other additional innovations used include pre-loading the input data to hide latency via \gls{cpu} \emph{pipelining}, compressing sparse activations into a \gls{csr} format on-the-fly, and keeping convolutional kernels in L2 cache. For more details see \citet{kurtz_inducing_2020}. 

We compare our \gls{cpu} timings for \gls{srigl} to DeepSparse in \cref{fig:deep-sparse-benchmark} and find similar latency; however, we note that DeepSparse is subject to a higher variability as evidenced by a larger standard deviation. Further, many of the innovations used to accelerate unstructured sparse networks with DeepSparse could equally be applied to networks trained with \gls{srigl}. 
\section{Comparison with Structured Pruning Methods}\label{sec:structured-pruning}
In the following table we compare several structured pruning methods to \gls{srigl}. The tabulated structured pruning methods typically prune and fine-tune a pretrained model, resulting in extended training duration compared to typical dense training. We report the inference \gls{flops}, top-1 accuracy, and number of epochs for each method in \cref{table:compare_structured_pruning}.
\begin{table}[b!]
\begin{center}
\begin{threeparttable}
\caption{\textbf{Top-1 ImageNet test accuracy of ResNet-50 for various structured pruning methods} compared with \gls{srigl} and Chase~\citep{yin_dynamic_2023}. All values, except for \gls{srigl}, are obtained from \citet{yin_dynamic_2023}.}\label{table:compare_structured_pruning}
\begin{tabular}{lccc}
\toprule
    Methods&Inference \gls{flops}&Top-1 Accuracy&Epochs\\ 
\midrule
    Uniform & 2.0G & 75.1\%  &300  \\
    Random & 2.0G  & 74.6\%  & 300 \\
     GBN~\citep{you2019gate} & 2.4G  & 76.2\%  &350\\
    LEGR~\citep{Chin_2020_CVPR} & 2.4G  & 75.7\%&  - \\
    FPGM~\citep{he2019filter}& 2.4G  & 75.6\%   & 200 \\ 
    TAS~\citep{dong2019network} & 2.3G  & 76.2\%  & 240 \\
    Hrank~\citep{lin2020hrank} &2.3G  & 75.0\%& 570 \\
    SCOP~\citep{tang2020scop} & 2.2G & 76.0\% & 230 \\
    CHIP~\citep{sui2021chip} & 2.2G & 76.3\%  & -\\
    Group Fisher~\citep{liu2021group} & 2.0G  & 76.4\%&  - \\
    AutoSlim~\citep{yu2019autoslim} & 2.0G  & 75.6\% &-  \\
    {CafeNet-R}~\citep{su2021locally} & 2.0G  & {76.5\%} &300 \\
    Chase-1\tnote{\textdagger}~\citep{yin_dynamic_2023}  &1.5G   & 76.6\%  &250 \\
    \gls{srigl}\tnote{\textdagger} & 2.0G & 74.7\% & 205 \\
    \gls{srigl}\tnote{\textdagger} & 2.0G & 76.2\% & 515 \\
\midrule
\midrule
    Uniform & 1.0G &  73.1\% & 300\\
    Random & 1.0G  & 72.2\% & 300 \\ 
    Group Fisher~\citep{liu2021group} & 1.0G  & 73.9\% & - \\
    {CafeNet-R}~\citep{su2021locally} & 1.0G  & {74.9\%} &300\\
    {CafeNet-E}~\citep{su2021locally} & 1.0G & {75.3\%} &300 \\
    Chase-2\tnote{\textdagger}~\citep{yin_dynamic_2023}  &0.9G  & 75.7\% &  250\\
    \gls{srigl}\tnote{\textdagger} & 1.0G & 71.5\% & 205 \\
    \gls{srigl}\tnote{\textdagger} & 1.0G & 73.6\% & 515 \\
\bottomrule
\end{tabular}
\begin{tablenotes}
\footnotesize
\item[\textdagger]~\gls{dst} methods. All other methods tabulated are structured pruning methods.
\end{tablenotes}
\end{threeparttable}
\end{center}
\end{table}
\end{document}